\documentclass[journal]{IEEEtran}

\usepackage{graphicx}
\usepackage{subfigure}
\usepackage{float}
\usepackage{amsmath}
\usepackage{booktabs}
\usepackage[ruled,vlined]{algorithm2e}
\usepackage{hyperref}
\usepackage{mathrsfs}

\usepackage{amssymb}
\usepackage{amsfonts}
\usepackage{color}
\usepackage{multirow}
\usepackage{marvosym}
\usepackage{caption}
\usepackage{footnote}
\usepackage{dutchcal}
\usepackage{ulem}
\usepackage{balance}

\begin{document}

\title{Revisiting Edge Perturbation for Graph Neural Network in Graph Data Augmentation and Attack}

\author{Xin Liu, Yuxiang Zhang, Meng Wu, Mingyu Yan,~\IEEEmembership{Member,~IEEE}, Kun He, Wei Yan, \\Shirui Pan,~\IEEEmembership{Sensor Member,~IEEE}, Xiaochun Ye, and Dongrui Fan,~\IEEEmembership{Sensor Member,~IEEE}
% <-this % stops a space
\thanks{X. Liu, Y. X. Zhang, M. Wu, M. Y. Yan, X. C. Ye and D. R. Fan are with the SKLP, Institute of Computing Technology, Chinese Academy of Sciences, Beijing 100086, and also with the University of Chinese Academy of Sciences, Beijing 100049, China. M. Y. Yan is the corresponding author (yanmingyu@ict.ac.cn).}%
\thanks{K. He is with the Renmin University of China, Beijing 100872, China.}
\thanks{W. Yan is with the SKLP, Institute of Computing Technology, Chinese Academy of Sciences, Beijing 100086, and also with the Zhongguancun Laboratory, Beijing, China.}
\thanks{S. R. Pan is with the Griffith University, Brisbane, Australia.}
}


%Zhongguancun Laboratory. yanwei@ict.ac.cn yanwei@mail.zgclab.edu.cn

\markboth{Journal of \LaTeX\ Class Files,~Vol.~14, No.~8, August~2015}%
{Shell \MakeLowercase{\textit{et al.}}: Bare Demo of IEEEtran.cls for IEEE Journals}


\maketitle
\begin{abstract}

Edge perturbation is a basic method to modify graph structures. It can be categorized into two veins based on their effects on the performance of graph neural networks (GNNs), i.e., graph data augmentation and attack. Surprisingly, both veins of edge perturbation methods employ the same operations, yet yield opposite effects on GNNs' accuracy. A distinct boundary between these methods in using edge perturbation has never been clearly defined. Consequently, inappropriate perturbations may lead to undesirable outcomes, necessitating precise adjustments to achieve desired effects. Therefore, questions of ``why edge perturbation has a two-faced effect?'' and ``what makes edge perturbation flexible and effective?'' still remain unanswered.

In this paper, we will answer these questions by proposing a unified formulation and establishing a clear boundary between two categories of edge perturbation methods. Specifically, we conduct experiments to elucidate the differences and similarities between these methods and theoretically unify the workflow of these methods by casting it to one optimization problem. Then, we devise Edge Priority Detector (EPD) to generate a novel priority metric, bridging these methods up in the workflow. Experiments show that EPD can make augmentation or attack flexibly and achieve comparable or superior performance to other counterparts with less time overhead.

\end{abstract}

%\begin{IEEEkeywords}
%Edge perturbation, graph data augmentation, graph data attack, graph neural network.
%\end{IEEEkeywords}

% \fi
\IEEEpeerreviewmaketitle

\section{Introduction}
%1. 图的随机性与复杂性，无处不在。现实中的很多数据都可以抽象为图，举个现实生活中的例子并说明在这张图中边的数量是海量的。举一些数据集的例子说明在大规模数据集中，边的数量可达上千万甚至上亿，而且远远超过节点的数量(给一个倍数关系)。
\IEEEPARstart{G}{raphs} are ubiquitous in the real world and contain a wealth of information. In many applied fields, data naturally exhibit a graph structure in which objects themselves and relationships between objects can be respectively abstracted as nodes and edges in a graph, for example, traffic networks \cite{traffic1,traffic2} in transportation and social networks \cite{social1,social2} in social informatics. Take the World Wide Web (WWW) as an example. Web pages and referencing relations between two pages are respectively regarded as nodes and edges in a web graph \cite{shi2018graph}. The modern WWW graph contains more than seven billion pages and one trillion URLs\footnote{https://www.worldwidewebsize.com}. In such a large-scale graph, edge is an important element whose amount far exceeds nodes and generally receives more attention.
%Given the vastness of graphs, one may think that edge perturbation within a certain degree in a graph will not affect the performance of learning representation from the graph. 
%Given the vastness of graphs, empirical speculation may hold the idea that edge perturbation within a certain extent will not affect the performance of learning representation from graphs.

%However, a wealth of research has been conducted to demonstrate that various edge perturbation methods can assuredly affect the performance of learning graph representations, for example, improving or declining the accuracy of graph-related tasks such as node classification and link prediction. 
Edge perturbation is a basic topology-level modification to change the structure of a graph and can further affect the performance of learning graph representations. For example, edge perturbation can be leveraged to improve the accuracy of graph-related tasks such as node classification and link prediction. Herein, we take graph neural network (GNN) \cite{scarselli2008graph}, a highly effective neural network model for learning graph representations, as an exemplar, and analyze the effect of edge perturbation on model performance. 
GNNs are widely used deep learning methods that combine efficient neural network models with graph learning \cite{survey1,survey2,survey3,survey4,survey5,liu2022survey,liu2021sampling,lin2022comprehensive,GPP_survey}, holding state-of-the-art performance on diverse graph-related tasks \cite{sota1,sota2,sota3,sota4,shao2022decoupled}.
Edge perturbation methods for GNNs typically add or remove edges to make perturbation. The perturbed graph is then fed to a GNN for learning, which will cause distinct variations in model accuracy on downstream tasks. Existing literature on edge perturbation can be divided into two veins according to their purposes, that is, graph data augmentation and attack (we use their abbreviations \textit{Gaug} and \textit{Gatk} in the rest of the paper).

\begin{figure}[t]
\centering
\includegraphics[width=0.59\columnwidth]{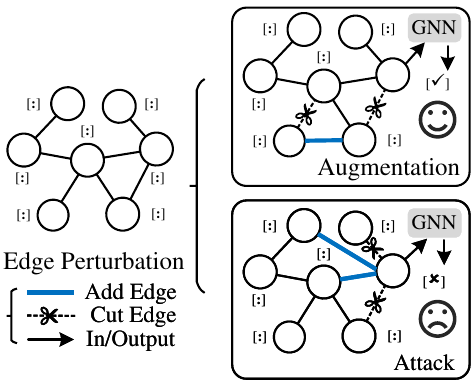}
\caption{Illustrations of edge perturbation methods in GNNs.} %Surprisingly enough, with different purposes, the same operation, i.e., perturbing edges in a graph, can yield opposite effects in terms of accuracy.}
\label{fig:teaser}
\end{figure}

The two veins of edge perturbation methods, surprisingly enough, have the same operations but bring about completely opposite influences with regard to GNNs' performance. 
As illustrated in Figure \ref{fig:teaser}, Gaug is a technique of wide scope used to improve the generalization of GNN models \cite{DAG_Survey1,DAG_Survey2,DAG_Survey3}. %Specifically, edge perturbation methods are topology-level augmentation techniques and are regarded as fundamental modifications to the graph structure \cite{DAG_Survey1}. 
Such methods propose to perturb edges in a graph to help alleviate the over-smoothing issue or improve the generalization \cite{dropedge}. By feeding the augmented graph, GNN generally gains better performance on classification or prediction than being trained on the original one.
Another vein of edge perturbation, i.e., Gatk, proposes to add or remove edges to generate adversarial attacks \cite{GraphAttack_survey} toward graph data and deteriorate GNNs' performance, which yields an opposite effect compared with Gaug. Such methods inject poison into an input graph, for example, they modify the graph topology to increase the misclassification ratio of GNNs via edge perturbation. Still, there has never been a clear boundary between Gaug and Gatk methods in using edge perturbation. One may obtain undesirable and even opposite performance when using a Gaug or Gatk method without precise adjustments.
%Therefore, we find edge perturbation methods for augmentation and attack have the same operation, i.e., perturbing edges in a graph, yet yield opposite effects in terms of accuracy. 
In this case, there comes up two open questions that ``\textit{why edge perturbation has a two-faced effect}'' and ``\textit{what makes edge perturbation flexible and effective}''.
%Dumped: Graph augmentation methods are able to improve the overall performance of classification or prediction on the graph, while graph attack methods ...

In this paper, we will answer the above questions through both theoretical analysis and experimental observations. We try to prove that Gaug and Gatk methods are essentially the same technique with different restricted conditions. To this end, we unify the edge perturbation methods for augmentation and attack, and establish a clear boundary between these methods by proposing a priority metric.
%This technique selectively perturbs edges to yield effects of augmentation and attack with the assistance of well-designed modules and strategies. 
Despite the opposite effect in accuracy, we find that Gaug and Gatk methods have similar operations and workflows, providing a potential opportunity to bridge them up.
%Specifically, we first conduct experiments using typical Gaug and Gatk methods to show an intriguing phenomenon. We train the same GNN model on two graph replicas generated from one graph under the same edge perturbation ratio and witness opposite influences on GNN's accuracy. 
Specifically, we conduct experiments on Gaug and Gatk methods to reveal their differences and similarities. Based on the similarities, we theoretically unify the workflow of Gaug and Gatk methods by casting it to an optimization problem. Then, we operationally unify the workflow of these methods to enable a flexible adjustment on making augmentation or attack by devising Edge Priority Detector (EPD). EPD can generate perturbation priorities of edges, based on which perturbations will be made for augmentation or attack flexibly. We indicate that perturbation priorities of edges will serve as the quantizable boundary to distinguish Gaug and Gatk methods. Based on the proposed priority metric, Gaug methods can remove edges of low importance to improve the model generalization, while Gatk methods can destruct key connections to corrupt accuracy. Moreover, thanks to the unified priority metric, EPD can make effective augmentation or attack with less time cost compared to other counterparts.

Contributions of this work can be summarized as follows.
\begin{itemize}
    %\item We reveal the reason for the opposite effect brought by graph data augmentation and attack methods. Typically, graph data augmentation methods are designed to eliminate redundant or task-irrelevant edges for promoting the model accuracy in corresponding tasks, while graph attack methods propose to remove important edges (e.g., semantic connections) or enlarge the loss of GNNs on poisoned graphs for declining the model accuracy in downstream tasks. We conduct extensive experiments and analyses with diverse graph datasets, GNN models, and edge perturbation methods. We also investigate the impact of the edge perturbation ratio on the effectiveness of various methods.

    \item We present the first work that systematically compares two categories of edge perturbation methods with similar operations yet completely opposite effects, i.e., Gaug and Gatk methods, from theoretical and experimental aspects.
    
    \item We unify Gaug and Gatk methods theoretically by proving that they share the same underlying technique and can be cast as one optimization problem, differing only in their constrained conditions. %The optimization problem is proven solvable.
    %We give a unified formulation for edge perturbation methods having opposite effects. Many graph data augmentation and attack methods transform the target of perturbing edges as an optimization problem, we thus bridge them up in a theoretical manner via a unified formulation.

    \item We devise Edge Priority Detector (EPD) to make augmentation or attack flexibly, by which we bridge Gaug and Gatk methods up in their workflows. EPD can make augmentation or attack flexibly and achieve comparable or superior performance to other counterparts with less time overhead. More importantly, EPD contributes a unified workflow and allows tailored adjustments in perturbation, which inspires bolder attempts in devising novel methods in this domain. %EPD can efficiently yield either effect of augmentation or attack. Instead of devising an elaborate Gaug or Gatk method, EPD contributes a unified workflow and allows tailored adjustments in perturbation, which inspires bolder attempts in devising novel methods in this domain.
    %We devise Edge Priority Detector (EPD), a plug-to-play module to make augmentation or attack to graphs based on priority metrics of edges, by which we operationally bridge Gaug and Gatk methods up in their workflows.
    %We devise Important Edge Detector (IED), a plug-to-play module to acquire the importance of edges in a graph. One can make augmentation or do damage to graphs based on acquired priority metrics in either scenario, by which we bridge graph augmentation and attack methods up in their workflows.

    \item We conduct a series of experiments to reveal correlations between the effectiveness of edge perturbation methods and various factors (e.g., graph attribute analysis in APPENDIX) and offer analysis of the augmentation and attack patterns and characteristics, aiming to contribute to the advancement of edge perturbation for augmentation and attack. %Code is provided open-source at \href{https://github.com/aiwen7/Awesome-Graph-Aug-Atk-via-Edge-Perturbation}{GitHub}.
    %We demonstrate the effectiveness of the proposed EPD. EPD can flexibly make augmentation and generate attacks by perturbing edges in a given graph. Moreover, a series of experiments are extensively given to make an in-depth analysis of edge perturbation methods. We reveal correlations between the effectiveness of edge perturbation methods and various factors (e.g., edge perturbation ratio, depth of GNNs, and graph attributes) and offer analysis of the augmentation and attack patterns, aiming to provide insightful suggestions for devising novel methods in this domain.
    
\end{itemize}

\section{Preliminary}
In this section, we first introduce fundamentals of GNNs and edge perturbation methods. Next, we put forward our motivation after conducting a preliminary experiment.

\subsection{Fundamentals of GNNs}
GNNs learn representations from complex graphs via a neural network model. Previous literature has fused widely used mechanisms in deep learning, such as convolution and attention operations, with plain GNNs for innovating various GNN variants, among which Graph Convolutional Networks (GCNs) \cite{gcn} are popularly applied to diverse scenarios. We take a vanilla GCN as an exemplar and formulize the procedures of graph learning. Given a graph $G (V, E)$, connections between nodes in the spatial dimension are represented by an adjacency matrix \textit{A}. Initial features corresponding to all nodes are represented by a feature matrix \textit{X} which will be updated layer by layer. Taking \textit{A} and \textit{X} as the input, GCN learns the hidden representation matrix \textit{H} in each layer \textit{l} in an iterative manner. The above process can be formulized as follows:
\begin{equation} \label{Eq:GCN}
    H^{l+1} = \sigma \left( \widetilde{D}^{-\frac{1}{2}} \widetilde{A} \widetilde{D}^{-\frac{1}{2}} H^{l} W^{l} \right)
\end{equation}
where $\widetilde{A}$ is a normalized adjacency matrix and $\widetilde{D}$ is a degree matrix of $\widetilde{A}$. $H^l$, $W^l$ are hidden feature and trainable weight in the $l$-th layer. And \textit{H} is initialized by \textit{X}, i.e., $H^0$ = \textit{X}. $\sigma(\cdot)$ denotes a nonlinear activation function such as ReLU. 
\iffalse
GAT similarly computes hidden representations in a layered manner. Differently, it calculates attention coefficients $\alpha_{ij}$ for node pairs (\textit{i, j}) during message propagation. 
\begin{equation} \label{Eq:GAT1}
    \alpha_{ij} = \frac{\mathrm{exp} \left( \mathrm{LeakyReLU}(\textbf{a}^{T}[\textbf{Wh}_{i} || \textbf{Wh}_{j}]) \right) }{\sum_{k \in N_i} \mathrm{exp} \left( \mathrm{LeakyReLU}(\textbf{a}^{T}[\textbf{Wh}_{i} || \textbf{Wh}_{k}]) \right) }
\end{equation}
where $\textbf{a}^{T}$ denotes a transposed weight vector, \textbf{W} denotes a weight matrix of the corresponding input linear transformation. \textbf{h$_{i}$} and $N_i$ denote the hidden feature and neighbor set of the node \textit{i}. Then, the output feature \textbf{h}$^{\prime}_{i}$ can be computed as follows:
\begin{equation} \label{Eq:GAT2}
    \textbf{h}_{i}^{\prime} = \sigma \left( \sum_{j \in N_i} \alpha_{ij} \textbf{W} \textbf{h}_{j} \right)
\end{equation}
where $\sigma(\cdot)$ denotes a nonlinear activation function, same as the Equation \eqref{Eq:GCN}. GAT also provides a multi-head attention mechanism \cite{attention} to improve the stability of learning. The computation of the output feature \textbf{h}$^{\prime}_{i}$ will be reformulated by adding extra concatenation and average operations.
\fi

\subsection{Fundamentals of Edge Perturbation}

\begin{table*}[t]
\centering
\caption{Summary of edge perturbation methods for GNNs. We remark that graph-related tasks, including node classification, graph classification, and link prediction, are abbreviated as NC, GC, and LP, respectively. Note$^1$: we only consider typical GNN models in this statistical item, meaning that conventional graph embedding methods are not involved.}
\label{tab:edge_perturbation_summary}
\resizebox{1\textwidth}{!}{
\begin{tabular*}{0.975\textwidth}{cccccc}  
\bottomrule
Category & Method & Venue & Approach & Applied GNN Model$^1$ & Task \\ \hline  
\multirow{9}{*}{\begin{tabular}[c]{@{}c@{}} Graph \\ Augmentation \end{tabular}}
& DropEdge\cite{dropedge} & ICLR$^{\prime}$20 & Remove Edges & GCN\cite{gcn}, GraphSAGE\cite{graphsage}, Deepgcns\cite{deepgcns}, ASGCN\cite{asgcn} & NC \\ \cline{2-6} 
& NeuralSparse\cite{aug-NeuralSparse} & ICML$^{\prime}20$ & Remove Edges & GCN\cite{gcn}, GraphSAGE\cite{graphsage}, GAT\cite{gat}, GIN\cite{gin} & NC \\ \cline{2-6}
& SGCN\cite{aug-sgcn} & PAKDD$^{\prime}$20 & Remove Edges & GCN\cite{gcn}, GraphSAGE\cite{graphsage} & NC \\ \cline{2-6}
& AdaptiveGCN\cite{aug-AdaptiveGCN} & CIKM$^{\prime}$21 & Remove Edges & GCN\cite{gcn}, GraphSAGE\cite{graphsage}, GIN\cite{gin} & NC \\ \cline{2-6}
& PTDNet\cite{aug-PTDNet} & WSDM$^{\prime}$21 & Remove Edges &  GCN\cite{gcn}, GraphSAGE\cite{graphsage}, GAT\cite{gat} & NC\&LP \\ \cline{2-6}
& TADropEdge\cite{aug-TADropEdge} & arXiv$^{\prime}21$ & Remove Edges & GCN\cite{gcn}, GIN\cite{gin} & NC\&GC \\ \cline{2-6}
& UGS\cite{aug-UGS} & ICML$^{\prime}$21 & Remove Edges & GCN\cite{gcn}, GAT\cite{gat}, GIN\cite{gin} & NC\&LP \\ \cline{2-6}
& GAUG\cite{aug-GAUG} & AAAI$^{\prime}$21 & Add/Remove Edges & GCN\cite{gcn}, GAT\cite{gat}, GraphSAGE\cite{graphsage}, JK-NET\cite{jk-net} & NC \\ \cline{2-6}
& AdaEdge\cite{aug-adaedge} & AAAI$^{\prime}$20 & Add/Remove Edges & GCN\cite{gcn}, GAT\cite{gat}, HG\cite{hypergraph}, GraphSAGE\cite{graphsage}, HO\cite{high-order} & NC \\ 

\hline

\multirow{11}{*}{Graph Attack}
& Nettack\cite{atk-nettack} & KDD$^{\prime}$18 & Add/Remove Edges & GCN\cite{gcn} & NC\\\cline{2-6} %CLN\DeepWalk
& Metattack\cite{atk-metattack} & ICLR$^{\prime}$19 & Add/Remove Edges & GCN\cite{gcn} & NC\\\cline{2-6} %CLN\DeepWalk
& LinLBP\cite{atk-linlbp} & CCS$^{\prime}$19 & Add/Remove Edges & GCN\cite{gcn} & NC\\ \cline{2-6} %RandomWalk node2vec均排除
& Black-box Attack\cite{atk-blackbox} & CCS$^{\prime}$21 & Add/Remove Edges & GIN\cite{gin}, SAG\cite{sag}, GUnet\cite{gunet} & GC \\ \cline{2-6} 
& Topo-attack\cite{atk-topology} & IJCAI$^{\prime}$19 & Add/Remove Edges & GCN\cite{gcn} & NC\\ \cline{2-6} 
& GF-attack\cite{atk-21} & AAAI$^{\prime}$20 & Add/Remove Edges & GCN\cite{gcn}, SGC\cite{sgc} & NC\\\cline{2-6}
& LowBlow\cite{atk-lowblow} & WSDM$^{\prime}$20 & Add/Remove Edges & GCN\cite{gcn} & NC \\ \cline{2-6} 
& GUA\cite{atk-GUA} & IJCAI$^{\prime}$21 & Add/Remove Edges & GCN\cite{gcn}, GAT\cite{gat} & NC \\ \cline{2-6} 
& NE-attack\cite{atk-NE-attack} & ICML$^{\prime}$19 & Add/Remove Edges & GCN\cite{gcn} & NC\&LP\\ \cline{2-6}
& Viking\cite{atk-VIKING} & PAKDD$^{\prime}$21 & Add/Remove Edges & GCN\cite{gcn} & NC\&LP\\ \cline{2-6}
& RL-attack\cite{atk-adversarial} & ICML$^{\prime}$18 & Add/Remove Edges & GCN\cite{gcn} & NC\&GC\\ 
\bottomrule
\end{tabular*}
}
\end{table*}

Edge perturbation methods modify the topology of an input graph \textit{G} by adding or removing edges from \textit{G}. The modification strategy can be specifically devised or follow a stochastic policy. From the topologic view, edge perturbation merely modifies the adjacency matrix extracted from \textit{G} while holding the amount and order of the node set of \textit{G} unchanged. The modified graph $G^{\prime}$ will be fed to a GNN model and exerts an influence on the model performance. Edge perturbation methods are de facto data-level operational approaches and will not modify the GNN model. Therefore, edge perturbation is flexible and highly compatible with various GNN models. Existing literature on edge perturbation can be divided into Gaug and Gatk according to their purposes. A detailed survey is given in Table \ref{tab:edge_perturbation_summary}. %To make an in-depth analysis, we further categorize each type of literature based on the mechanism of making perturbation.
%\noindent \textbf{Categorization for Graph Data Augmentation Methods.} \noindent \textbf{Categorization for Graph Attack Methods.}

To enable augmentation on a graph, previous literature \cite{aug-adaedge,aug-AdaptiveGCN,aug-GAUG,aug-NeuralSparse,aug-PTDNet,aug-sgcn,aug-TADropEdge,aug-UGS} perturbs edges in an input graph by adding or removing connections between nodes. The majority of these methods use a sparsification strategy to convert the input graph to a sparser one. The key idea of these methods consists in how to devise a suitable sparsifier. Typically, 
DropEdge \cite{dropedge} pioneers to randomly remove a fixed ratio of edges in each training epoch, which aims to resolve the over-smoothing issue in deep GNN models.
NeuralSparse \cite{aug-NeuralSparse} utilizes a Multilayer Perception (MLP) based model to learn a sparsification strategy based on the feedback of downstream tasks in training.
SGCN \cite{aug-sgcn} casts graph sparsification as an optimization problem and resolves the problem via the alternating direction method of multipliers (ADMM) method \cite{ADMM}.
UGS \cite{aug-UGS} utilizes the lottery ticket hypothesis \cite{lth} to sparsify input graphs and a GNN model iteratively. 
PTDNet \cite{aug-PTDNet} removes task-irrelevant edges and applies a nuclear norm regularization to impose a low-rank constraint on the sparsified graph.
AdaptiveGCN \cite{aug-AdaptiveGCN} removes task-irrelevant edges based on the feedback of downstream tasks acquired by an edge predictor module.
TADropEdge \cite{aug-TADropEdge} computes weights for edges in a graph based on the graph spectrum and adaptively removes edges according to normalized weights.
Instead of merely performing sparsification, GAUG \cite{aug-GAUG} and AdaEdge \cite{aug-adaedge} adjust the graph topology by dynamically adding and removing edges based on the predicted metrics. The metrics are related to edges and are acquired based on an edge predictor and model prediction accuracy, respectively.

To generate attacks on a graph, previous literature \cite{atk-21,atk-adversarial,atk-blackbox,atk-linlbp,atk-lowblow,atk-metattack,atk-nettack,atk-topology,atk-NE-attack,atk-VIKING} utilizes various methods to modify the topology of an input graph, resulting in the corruption of GNN models in terms of accuracy. The central idea behind these methods is to identify and selectively perturb key edges in a graph, thus rendering a significant performance decline in node and subgraph classification. For examples,
Nettack \cite{atk-nettack} applies structural perturbations and node-level feature perturbations to make the target nodes easy to be misclassified.
Metattack \cite{atk-metattack} treats the graph structure as an object to be optimized. It utilizes the optimization results to modify the adjacency matrix, thus injecting perturbations to the input graph.
Besides, many studies \cite{atk-linlbp,atk-blackbox,atk-topology,atk-VIKING} cast attacking graph structure as optimization problems. LinLBP \cite{atk-linlbp} proposes to minimize the total cost of modifying the graph structure while maximizing the False Negative Rate (FNR). There are constraints imposed on the FNR values around the target node and the total number of edge additions or deletions between pairs of nodes. Black-box Attack \cite{atk-blackbox} and GF-attack \cite{atk-21} focus on generating hard label black-box attacks, where an attacker has no knowledge about the target GNN model and can only obtain predicted labels through querying the target model.
Viking \cite{atk-VIKING} proposes a bi-level optimization to generate attacks on graphs and approximately addresses the optimization problem via a brute-force solution.
When optimization-based attacks fail to effectively solve the problem, Topo-attack \cite{atk-topology} proposes gradient-based attacks as an alternative approach.
Other methods, such as LowBlow \cite{atk-lowblow} reveals that attacks often exhibit low-rank characteristics. 
GUA \cite{atk-GUA} finds that anchor nodes obtained from a small training set can attack a majority of target nodes in classification tasks.
NE-attack \cite{atk-NE-attack} demonstrates that node embeddings are vulnerable under topology attacks.
RL-attack \cite{atk-adversarial} learns a generalizable attack policy via reinforcement learning, and applies genetic algorithms and gradient descent to different attack scenarios.

\subsection{Motivation}
% 建议在三个数据集cora citeseer pubmed上分别是测试一个两层的基础GCN分别使用DropEdge[https://github.com/DropEdge/DropEdge]和Topo-attack[https://github.com/KaidiXu/GCN_ADV_Train]，在相同扰动率的条件下，模型精度。相同扰动率：
%DropEdge中/src/train_new.py/sampling_percent 参数
%Topo-attack中/attack.py/perturb_ratio 参数
%DropEdge Dataset = cora, PR = 0.05, Acc = 0.8620
%DropEdge Dataset = cora, PR = 0.1, Acc = 0.8790
%DropEdge Dataset = cora, PR = 0.3, Acc = 0.8710
%DropEdge Dataset = citeseer, PR = 0.05, Acc = 0.7890
%DropEdge Dataset = citeseer, PR = 0.1, Acc = 0.7920
%DropEdge Dataset = citeseer, PR = 0.3, Acc = NA
%DropEdge Dataset = pubmed, PR = 0.05, Acc = 0.9020
%DropEdge Dataset = pubmed, PR = 0.1, Acc = 0.9040
%DropEdge Dataset = pubmed, PR = 0.3, Acc = NA
%Topo-attack Dataset = cora, PR = 0.05， Acc = 0.73200
%Topo-attack Dataset = cora, PR = 0.1, Acc = 0.69200
%Topo-attack Dataset = cora, PR = 0.3, Acc = 0.55100 
%Topo-attack Dataset = citeseer, PR = 0.05， Acc = 0.64900
%Topo-attack Dataset = citeseer, PR = 0.1, Acc = 0.60500
%Topo-attack Dataset = citeseer, PR = 0.3, Acc = 0.47900 
%Topo-attack Dataset = pubmed, PR = 0.05, Acc = 0.56800
%Topo-attack Dataset = pubmed, PR = 0.1, Acc = 0.44800
%Topo-attack Dataset = pubmed, PR = 0.3, Acc =  0.33300

Two categories of edge perturbation methods have been previously introduced to reveal their characteristics and opposite influences on model accuracy. Next up, we conduct a preliminary experiment to show an intriguing phenomenon and put forward our motivation. Taking a two-layer GCN \cite{gcn} as the backbone model, we select two Gaug and Gatk methods, i.e., UGS \cite{aug-UGS} and Viking \cite{atk-VIKING}, to perturb edges. We use the methods to generate perturbations on three real-world graphs, with the same model-specific parameters leveraged during training GCN. By changing the edge perturbation ratio (EPR, a quantity ratio of the perturbed edges to the total edges), we record varying performances of GCN on Cora, Citeseer, and Pubmed datasets (abbr. CR, CS, PD). 

\begin{figure}[t]
\centering
\includegraphics[width=0.85\columnwidth]{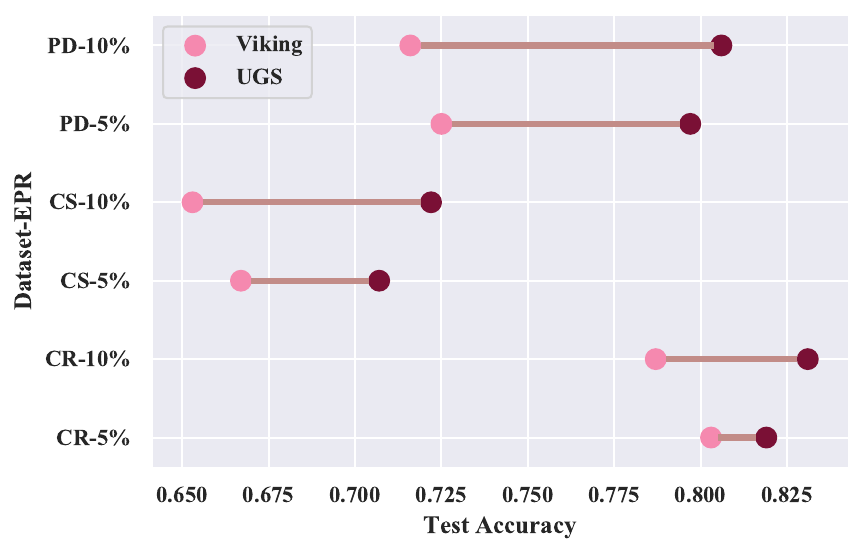}
\caption{Variations of test accuracy when applying Gaug and Gatk methods on three datasets under different EPR.}
\label{fig:motivation}
\end{figure}

As illustrated in Figure \ref{fig:motivation}, in the case where configurations of GCN, input graph, and EPR are the same, the accuracy of GCN with UGS (augmentation) applied far exceeds that with Viking (attack) applied. The accuracy gap expands if we increase EPR within a certain range.
The above experimental observations raise the question of ``\textit{why edge perturbation has a two-faced effect}''. 
%We \textbf{answer} that different edge perturbation methods can offer opposite influences with respect to GNN performance, according to the design goals. Either in augmentation or attack, edge perturbation is effective in its applied scenarios. 
We answer that different edge perturbation methods will perturb different edges to achieve their purposes. Gaug methods remove edges of low importance to improve the model generalization and promote accuracy, while Gatk methods destroy key edges to corrupt the topology of a graph and cause accuracy loss.
The above answer further elicits a question that ``\textit{what makes edge perturbation flexible and effective}''.

%To answer the question, we put forward our viewpoints as follows. To begin with, graph augmentation and attack methods are all edge perturbation methods with distinct purposes. These methods modify the graph topology to affect the GNN performance on learning representations from the perturbed graph. We argue that artificial-guided edge perturbations can remove redundant information to alleviate over-smoothing or destroy the integrality of a graph to cause incorrect classification and prediction, depending on our goals. Thus, the critical matter in edge perturbation methods is whether key connections (in other words, important edges) in a graph are perturbed, which directly determines the type of the caused influence. Specifically, graph augmentation methods tend to preserve important edges and remove redundant edges, while graph attack methods aim to break important edges. Accordingly, we derive two conclusions from the above discussions: 
To answer the question, %we put forward our viewpoints as follows. %To begin with, Gaug and Gatk methods are all edge perturbation methods with distinct purposes. These methods modify the graph topology to affect the GNN performance on learning representations from the perturbed graph. 
we argue that artificial-guided edge perturbations can improve the model generalization or destroy the integrality of a graph, under the goal of augmentation or attack. Thus, the critical matter in edge perturbation methods is how to find a boundary between Gaug and Gatk methods, in other words, which edges are perturbed in priority, under a given goal. Accordingly, we can derive two conclusions from the above discussions: 

1) The \textit{flexibility} of edge perturbation methods attributes to the fact that they can be uniformly formalized and performed.

2) The \textit{effectiveness} of edge perturbation methods comes from their capability of detecting perturbation priorities of edges to achieve either effect of augmentation or attack.

Therefore, in Section \ref{Sec:3}, we pioneer unifying Gaug and Gatk methods by casting them as an optimization problem with two independent constraints. Moreover, we propose to devise a novel module to acquire the perturbation priority to enable flexible augmentation and attack. 

\begin{figure}[t]
\centering
\includegraphics[width=0.8\columnwidth]{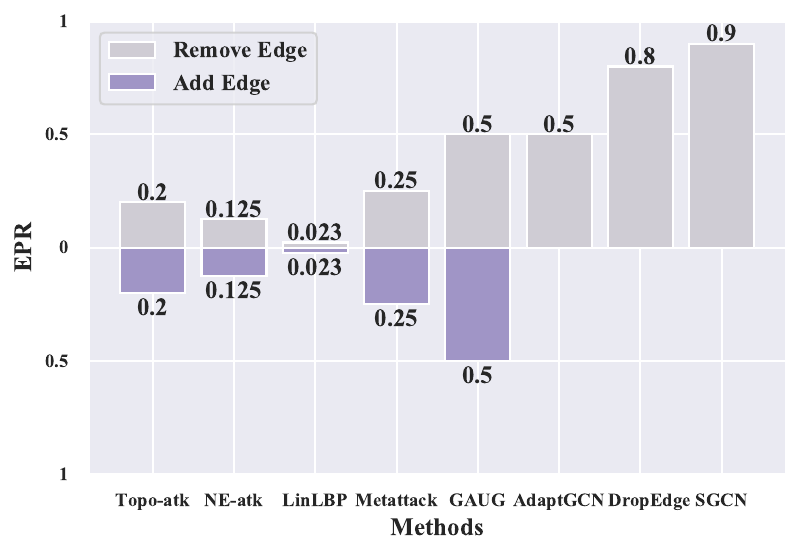}
\caption{Statistics on the maximal reported EPR among typical methods. The first four from left to right are Gatk methods, and the rest are Gaug methods.}
\label{fig:compare}
\end{figure}

%The above experimental results raise an open question, i.e., ``whether edge perturbation is beneficial to GNNs''. To answer the question, we put forward our viewpoints as follows. To begin with, graph augmentation and attack methods are all edge perturbation methods with distinct purposes. These methods modify the graph topology to affect the GNN performance on learning representations from the perturbed graph. We argue that artificial-guided edge perturbations can remove redundant information to alleviate over-smoothing or destroy the integrality of a graph to cause incorrect classification and prediction, depending on our goals. Thus, the critical matter in edge perturbation methods is whether key connections (in other words, important edges) in a graph are perturbed. Supposing the matter is resolved, graph augmentation and attack methods will be bridged together through a unified formulation. In this paper, we pioneer unifying graph augmentation and attack methods by casting them as an optimization problem with two independent constraints. 

 \section{Methodology} \label{Sec:3}
In this section, we aim to unify Gaug and Gatk methods by proving that they are essentially the same technique with different restricted conditions. First, we revisit Gaug and Gatk methods and rethink their similarities and differences. Next, we put forward a unified formulation to bridge these methods together theoretically. Then, we devise a module to acquire priority metrics of edges to establish a quantizable boundary for making flexible perturbations.

\subsection{Revisiting Edge Perturbation} \label{Sec:3.1}

\begin{figure*}[t]
\centering
\includegraphics[width=0.83\textwidth]{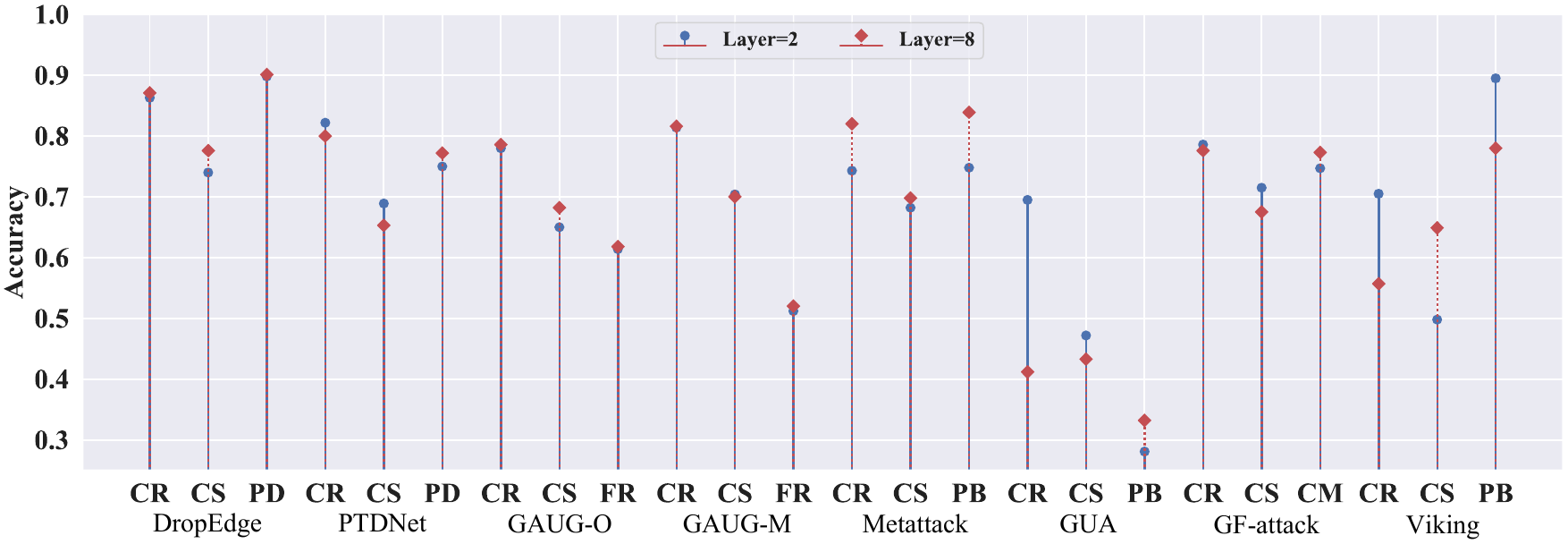}
\caption{Comparisons of the accuracy of GNN models (2-layer GNN and 8-layer GNN) with different Gaug and Gatk methods applied. For each dataset, the only variate is the number of layers of GNN models as we keep all other parameters regarding GNN models and training configurations the same, including GNN backbones, learning rate, EPR, etc.}
\label{fig:DIFFinDeepModel}
\end{figure*}

\iffalse
...we unify the metric to accuracy. In the context of adversarial attacks on graphs, accuracy serves as the unified performance metric. However, different Gatk methods employ varying approaches to compute accuracy. Specifically, in the case of Metattack\cite{atk-metattack}, they train and employ a meta-learning approach to attack surrogate Graph Convolutional Network (GCN\cite{gcn}) models, measuring the model's output accuracy. In GFAttack\cite{atk-21}, they perform an analysis of the graph data's topological structure to identify a set of potentially disruptive edges with the highest scores. By perturbing these edges, they construct poisoned training and testing sets, leading to a reduction in the output accuracy of the GCN-based surrogate model. Similarly, GUA\cite{atk-GUA} generates perturbations on the graph edges and evaluates their impact on a baseline scenario. In the Viking\cite{atk-VIKING} method, an optimal set of edges susceptible to attacks is identified, and the accuracy of the GCN model is diminished following perturbation. Across these attack methodologies, the downstream task associated with GCN is node classification. The accuracy decreases when the labels assigned to nodes by the GCN do not align with their correct labels. Further comprehensible is the notion that GUA\cite{atk-GUA} designates victim classes, denoting the specific class that, according to their attack methodology, nodes are more likely to be misclassified into.
\fi

\textit{1) Rethinking the Differences: Extent of Perturbation}

Differences between Gaug and Gatk methods lie in the extent of making perturbations on a graph, except for distinctions in mechanisms. As illustrated in Figure \ref{fig:compare}, we made statistics on the maximal EPR of typical Gaug and Gatk methods. We can derive two findings.
\textit{First}, most Gaug methods make perturbations by removing edges in a graph to boost generalization. Making augmentation by adding and removing edges relies heavily on a well-designed predictor module \cite{aug-GAUG,aug-adaedge}. Recent literature has investigated that randomly adding edges, similar to the mechanism of DropEdge, can not improve the accuracy of the node classification task in most cases \cite{SunLYFPJLY22}. Whereas Gatk methods can flexibly add and remove edges to inject attacks into a graph, for example, Viking \cite{atk-VIKING} proposes adding and removing edges to make nodes' features similar, thus leading to misclassification in downstream tasks.
\textit{Second}, most Gatk methods propose to keep the perturbation ratio within a certain extent, in other words, they recommend making the perturbations hard to be noticed. For example, Black-box Attack \cite{atk-blackbox} and Topo-attack \cite{atk-topology} cast generate attacks as an optimization problem where constraint conditions consist of an upper bound of the edge perturbation ratio. Whereas Gaug methods have no restrictions on the edge perturbation ratio. Many methods try to expand the extent of adding perturbations to seek an empirical optimal ratio. For example, SGCN \cite{aug-sgcn} incorporates the edge perturbation ratio into the constraint conditions of an optimization problem by setting a low bound. SGCN expects to derive a sparse enough graph $G^\prime$ and achieve comparable accuracy when it feeds $G^\prime$ to a trained GCN. 
%To summarize, Gatk methods propose making minor perturbations to generate attacks that can cause a significant decline in accuracy. The perturbation types include adding and removing edges, in other words, flipping edges. However, Gaug methods make perturbations by removing edges mostly, yet do not give a restriction to the perturbation ratio.

\textit{2) Rethinking the Differences: Effect on Deep GNNs} 

The difference between Gaug and Gatk methods is also reflected in the effect of augmentation/attack varies with the depth of GNN models. Previously, the authors of GCN \cite{gcn} have indicated that stacking the graph convolutional layer to construct a deep model generally can not bring about a better performance than a shallow one. This is because the performance of a deep GNN is impaired by several issues, among which over-smoothing is a critical reason causing performance decline \cite{oversmoothing1,oversmoothing2}. 
A typical Gaug method, DropEdge \cite{dropedge}, has proposed that the over-smoothing issue can be alleviated by randomly removing edges in a graph. %Other literature, for example, GAUG \cite{aug-GAUG}, TADropEdge \cite{aug-TADropEdge}, and PTDNet \cite{aug-PTDNet}, have also studied the effect of their proposed edge perturbation methods in deep GNNs. While, there has been few studies regarding the effect of Gatk methods under deep GNNs. 
Literature \cite{AirGNN} has found that the performance of a deep GCN is better than a shallow one in abnormal node classification. The abnormal nodes are defined as nodes whose features are naturally noise-involved or adversarially manipulated \cite{AirGNN}. Herein, Gatk methods perturb edges and modify the topology of a graph to generate abnormal connections among node regions. Thus, it is intriguing to observe what effect a deep GNN can have on Gatk methods. 

As illustrated in Figure \ref{fig:DIFFinDeepModel}, we evaluate the accuracy of GNN models with different augmentation/attack methods applied. Please note that many Gatk methods release the misclassification ratio to demonstrate the effectiveness of their generated attacks. Thereby, we convert the reported misclassification ratio to an accuracy metric to unify the evaluation metric for Gaug and Gatk methods. For instance, we count the number of correctly classified nodes by subtracting the misclassified and attacked (resulting in misclassification) nodes from the overall test set, then we calculate the classification accuracy as the evaluation metric for Gatk methods. For augmentation, we adopt DropEdge \cite{dropedge}, PTDNet \cite{aug-PTDNet}, and two variant models of GAUG \cite{aug-GAUG} to make augmentations on Cora, Citeseer, Pubmed, and Flickr (abbr. CR, CS, PD, FR) datasets. In most cases, we find deep GNNs with Gaug methods adopted generally outperform shallow GNNs in accuracy, despite the performance of deep GNNs being impaired by over-smoothing. For attack, we adopt Metattack \cite{atk-metattack}, GUA \cite{atk-GUA}, GF-attack \cite{atk-21}, and Viking \cite{atk-VIKING} to generate attacks on Cora, Citeseer, PolBlogs, and Cora-ML (abbr. CR, CS, PB, CM) datasets. In almost half of the cases, the accuracy of a deep GNN is higher than that of a shallow one when the same Gatk method is applied. Especially for Metattack, an 8-layer GCN is superior to a 2-layer GCN in classification with the same attack (same EPR) injected. We also observe variations between the accuracy of 2-layer and 8-layer GCNs are tiny with GUA and GF-attack applied, except for the case that the accuracy of the deep model is significantly lower than that of the shallow model. Given the fact that the accuracy of a deep GNN is generally lower than that of a shallow GNN, we can derive a conclusion that is similar to the findings in literature \cite{AirGNN}, i.e., a deep GNN may have a better ability than a shallow one in classifying the attacked nodes. This conclusion implies that some Gatk methods (e.g., Metattack) may gradually lose efficacy as the number of model layers deepens. 

\textit{3) Rethinking the Similarities: Workflow}

Similarities between Gaug and Gatk methods lie in their workflows. Generally, edge perturbations modify the topology of a graph by adding and removing edges. The operation of making perturbations is the same for Gaug and Gatk methods. We can formulate the operation from a mathematical aspect.
Given a graph \textit{G(V, E)} as the input, the adjacency matrix $A$ can be fetched by extracting the connections among nodes. Edge perturbation methods modify the topology of \textit{G} by perturbing connections in \textit{A}. Since \textit{A} is a sparse binary matrix, we define a binary format perturbation matrix as \textit{P} = [$p_{ij}$] whose element $p_{ij}$ is given by $p_{ij} \in$ \{0,1\}. $p_{ij}$ = 1 denotes edge perturbation will be made in the corresponding location of \textit{A}. The edge perturbation operation indicates adding a new edge or removing an existing edge in a specific location of \textit{A}. $p_{ij}$ = 0 indicates no perturbation will be made in the corresponding location. Thus, the perturbed adjacency matrix $A$ can be formulated as:
\begin{equation} \label{Eq:PerturbeADJ}
    A = A_0 \oplus P
\end{equation}
where $\oplus$ is an element-wise XOR operation. $A_0$ denotes the initial adjacency matrix. \textit{P} is of the same shape as \textit{A}. 

\subsection{Unified Formulation: Optimization Problem and Solution}
%2. 从顶层进行抽象，参照2.1：A' = A XOR P作为统一的核心公式引出讨论。对于基于优化问题的两种场景，将两种方法用一个统一过程进行表述
%2.1 图攻击情境中, 现有的方法大多用两个方法(待考证)生成攻击数据: 通过生成具有对抗攻击效果的训练数据来增大模型误分类的概率; 将生成对抗图数据转换为一种优化问题来求解, 生成的带污染的数据集被用于训练GNN
%2.2 图攻击转换的优化问题求解, 优化函数的目标一般是最小化攻击成本, 同时附带两大类型的约束: 对邻接矩阵扰动尽可能小(添加攻击前后, 使邻接矩阵差的范数越小), 误分类效果更好(对于同一个代理模型, 添加攻击前后节点被分类的标签不同，即出现分类错误)
%2.3 图数据增强的优化问题求解, 和图攻击场景下的优化问题是否具有共性
% SGCN: A Graph Sparsifier Based on Graph Convolutional Networks; Robust Graph Representation Learning via Neural Sparsification (An approximation function in Sec.IV)

%unified formulation: 优化问题表述为产生新的邻接矩阵或者改名字的邻接矩阵，关键在于不断迭代，迭代的过程就是产生A'的过程，迭代的结果就是新A'。所以优化函数的形式可以是：搞一个扰动矩阵S，A和S取异或的结果记为A‘
\iffalse 
\begin{equation}
    \begin{aligned}
    A^{'} = A\quad {\small xor} \arg \max_{S} &\sum S_{ij} C^{'}_{ij} , i,j \in S \\
    s.t. \sum_{i,j}\overline{C}^{(t)} & > \sum_{i,j}\overline{C}^{(t-1)} \\
    \end{aligned}
\end{equation}
\fi
%式子中，A是不定的，C是定的，C_{ij}即代表EPD所探测到的ij节点间的重要性，其角标(t)代表t轮迭代时的值。最终的目标是获得一个邻接矩阵，使得这个矩阵中每对有关系的节点的边重要性之和最大。f记为Loss函数，公式写的时候可以这么写，但是解的时候这样肯定没法解。有论文提到需要先对目标函数进行处理，首先就是C是个0-1矩阵，需要弛豫到[0,1]范围内变离散为连续，解完之后再离散回来。

\begin{figure*}[t]
\centering
\includegraphics[width=0.85\linewidth]{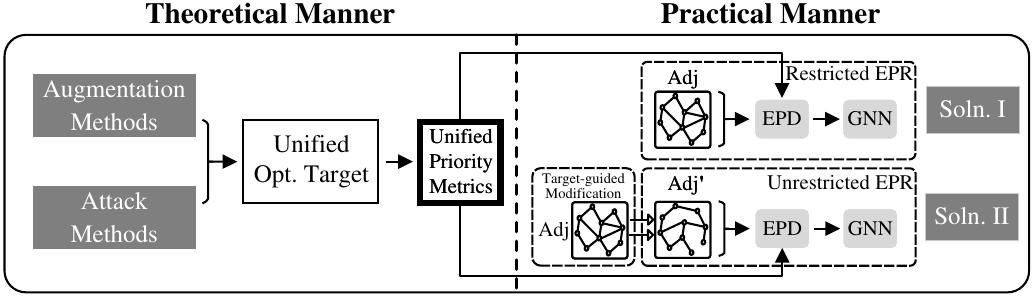}
\caption{Gaug and Gatk methods are bridged theoretically and practically by proposing unified priority metrics. The metrics will be leveraged in EPD to efficiently perturb edges, enabling flexible effects of attack or augmentation.}
\label{fig:EPD}
\end{figure*}

As previously discussed, the similarity between Gaug and Gatk methods is to derive a perturbed adjacency matrix \textit{A} for subsequent training. Herein, we propose to provide a unified formulation for Gaug and Gatk methods by casting the workflows of finding \textit{A} as an optimization problem. The unified formulation will be strong evidence showing that edge perturbation can \textit{flexibly} produce different effects (augmentation or attack). Given an input graph \textit{G(V, E)}, the original adjacency matrix of \textit{G} is defined as $A_0$. 
Let a well-trained GNN model be $g(A)$. Given a node $v_i$, the forward output of the model is $z_i = g(A, v_i)$. The label of the node $v_i$ can be fetched as $y_i = h(v_i)$. A general loss function of the model can be defined in a cross-entropy form:
\begin{equation} \label{Eq:Loss}
    L(A) = - \sum \limits_{v_i \in V} \sum \limits_{j} y_j ln \big( g(A, v_i)  \big)
\end{equation}
Based on Equation \eqref{Eq:Loss}, we can formulate making edge perturbation for an input graph \textit{G} (adjacency matrix \textit{A}) as the following optimization problem:
\begin{equation} \label{Eq:Opt}
    A^* = \mathop{\arg \min} \limits_{A} L(A)
\end{equation}
%\vspace{-2mm}
\begin{equation*}
    \begin{split}
        s. t. \left \{
                \begin{array}{ll}
                    ||A-A_0||_{0} = b_{aug}, y_j = h(v_i), \textrm{ for augmentation} \\
                    ||A-A_0||_{0} = b_{atk}, y_j \neq h(v_i), \textrm{ for attack}
                \end{array}
            \right.
    \end{split}
\end{equation*}
where \textit{b} denotes the budget of perturbed edges. The value of \textit{b} is distinct in augmentation and attack cases, and will be set to an even number since edge perturbations come in pairs in an adjacency matrix (symmetry of \textit{A}). In Equation \eqref{Eq:Opt}, we give two independent constraint conditions in the optimization problem to respectively suit the cases of Gaug and Gatk methods. 

\iffalse
\begin{equation}
    \begin{split}
        I(A) = \left \{
                \begin{array}{ll}
                    -N, y_j = h(v_i), \textrm{ for augmentation} \\
                    +N, y_j \neq h(v_i), \textrm{ for attack}
                \end{array}
            \right.
    \end{split}
\end{equation}
\begin{equation} \label{Eq:Opt}
    \begin{aligned}
        A^* = \mathop{\arg \min} \limits_{A} L(A) + I(A) \\
        s.t. \quad ||A-A_0||_{0} = budget
    \end{aligned}
\end{equation}
\fi

Specifically, for augmentation, it is expected to minimize the cross-entropy loss, thus promoting the probability of classifying a node to the real label. In this case, the label \textit{y} of a given node \textit{v} can be easily fetched via a label lookup function $h(\cdot)$.
The majority of Gaug methods \cite{dropedge,aug-NeuralSparse,aug-sgcn,aug-AdaptiveGCN,aug-PTDNet,aug-TADropEdge,aug-UGS} propose to remove edges for making perturbations. In some cases \cite{dropedge,aug-sgcn}, a significant extent of edges (more than 50\%, please see Figure \ref{fig:compare}) is removed from the original graph. To suit the requirement, $b_{aug}$ can be set to a large value to make the perturbed adjacency matrix $A$ sparse enough.
For attack, it is expected to corrupt a GNN model by increasing the probability of misclassification. Inspired by the literature \cite{CW-attack} targeting generating adversarial attacks from an optimization aspect, we inject attacks by inducing nodes to be misclassified. The label \textit{y} will be set to be the wrong label of a given node \textit{v} in the objective function. Besides, the setting of budget $b_{atk}$ is different from that in the augmentation case. Previous literature \cite{atk-blackbox,atk-linlbp,atk-topology} argues that it would be better to make attacks unobservable. In this case, setting \textit{b} to a small value is suitable for reducing the difference between the perturbed adjacency matrix and the original one. 

%\subsection{Unified Formulation: Solution to the Optimization Problem}
The optimization problem in Equation \eqref{Eq:Opt} is intractable to solve, since $L_0$ norm in the constraint condition is relevant to the adjacency matrix that is in a binary format. $L_0$ norm is non-differentiable and non-convex and has been an NP-hard problem. Therefore, we first replace $L_0$ norm with $L_1$ norm.

\noindent Proposition 1. \textit{Given two binary matrices A and B, the values of $L_0$ norm and $L_1$ norm of (A-B) are identical.} \\
\noindent Let $C = A - B$, $C = [c_{ij}] \in \{-1, 0, 1\}$. The $L_0$ norm counts the number of non-zero elements in \textit{C}, while the $L_1$ norm sums the absolute values of the elements in \textit{C}. In this case, we have $||C||_0$ equal to $||C||_1$. Next, we reformulate the optimization problem in Equation \eqref{Eq:Opt} via augmented Lagrangian \cite{aug-lagrangian}. The augmented Lagrangian of the problem is given as follows:
\begin{small}
\begin{equation} \label{Eq:Opt-Lagrangian}
    F(A, \lambda) = L(A) + \lambda^{T} \big( ||A-A_0||_1-b \big) + \frac{\rho}{2} ||\big( ||A-A_0||_1-b \big)||_{2}^{2} 
\end{equation}
\end{small}
where $\lambda$ is the Lagrangian multiplier, $\rho$ is a penalty parameter. We try to update variables \textit{A} and $\lambda$ iteratively via gradient descent. Whereas, given that \textit{A} is originally in binary format, updating it may modify elements in \textit{A} to floating point numbers. We conduct a bounded projection to the adjacency matrix and transform elements in it to continuous values ranging from 0 to 1, for example, using Sigmoid function for value projection. Note that the projection should be applied to both \textit{A} and $A_0$ in the initialization stage. Then, we can derive the projected adjacency matrix as $A = [a_{ij}] \in [0, 1]$. Each element $a_{ij}$ in $A$ will be regarded as the probability of making perturbations in the corresponding location. In this case, if a perturbed adjacency matrix $A^{\prime}$ is fetched by resolving the optimization problem, we can make perturbations based on:
\begin{small}
\begin{equation} \label{Eq:perturbation}
    \begin{split}
        A^{\prime} = [a^{\prime}_{ij}] = 
            \left \{
                \begin{array}{rr}
                    \textrm{perturb edges at (i, j) location, } \text{if } a^{\prime}_{ij} \geq \zeta \\
                    \textrm{no perturbation will be made, } \text{if } a^{\prime}_{ij} < \zeta
                \end{array}
            \right.
    \end{split}
\end{equation}
\end{small}
where $\zeta$ is a pre-set threshold parameter to control the action of making perturbations. Thereby, the iterative update formulas of variables \textit{A} and $\lambda$ based on augmented Lagrangian are:
\begin{equation} \label{Eq:update}
    \begin{aligned}
        A^{k+1} &= \mathop{\arg \min} \limits_{A} F(A, \lambda^{k}) \\
        \lambda^{k+1} &= \lambda^{k} + \rho(||A-A_0||_1-b)
    \end{aligned}
\end{equation}
In Equation \eqref{Eq:update}, $L_1$ norm is involved in the objective function, which will introduce non-differentiable points in gradient descent. To solve the above problem, we first rewrite the objective function in Equation \eqref{Eq:update} as:
\begin{equation} \label{Eq:RewriteObj}
    \begin{aligned}
        A^{k+1} &= \mathop{\arg \min} \limits_{A} f(A) + \tau (A) \\
        f(A) &= L(A) + \frac{\rho}{2} ||\big( ||A-A_0||_1-b \big)||_{2}^{2}
    \end{aligned}
\end{equation}
\textit{f(A)} is a differentiable function as cross-entropy loss and $L_2$ norm functions are all differentiable. $\tau(A)$ contains $L_1$ norm and is non-differentiable. We can utilize Proximal Gradient Descent (PGD) \cite{boyd2004convex} to solve the problem and derive \textit{A} as:
\begin{small}
\begin{equation} \label{Eq:PGD}
    \begin{aligned}
        A^{k+1} = \textrm{Prox}_{\tau, \eta} \big(A^k &- \eta \nabla f(A^k) \big) \\
        \textrm{Prox}_{\tau,\eta}(A) = \mathop{\arg \min} \limits_{\Gamma} \frac{||\Gamma - A||^2_2}{2}  &+ \eta(||\Gamma - A_0||_1 - b) = \hat{\Gamma}
    \end{aligned}
\end{equation}
\end{small}
with 
\begin{equation*} \label{Eq:ThreeCases}
    \hat{\Gamma}_{ij} = 
    \begin{cases}
    a_{ij} - \eta & \text{if}\ a_{ij}-\eta > 0, \\
    a_{ij} + \eta & \text{if}\ a_{ij}+\eta < 0, \\ 
    0 & \text{otherwise}, \\ 
    \end{cases}
\end{equation*}
where $\eta$ is a step size and $\textrm{Prox}_{\tau, \eta}(\cdot)$ denotes a proximal mapping of the function $\tau(\cdot)$. 

So far, we have shown that Gaug and Gatk methods are similar in their workflows and can be uniformly formulated from an optimization perspective. We have proved the unified formulation solvable via the steps above, thus indicating that Gaug and Gatk methods are essentially the same technique with different restricted conditions. 
Inspired by the formulation that Gatk and Gaug methods can be cast as a problem with a unified optimization target, we argue that a unified metric can be used in either attack or augmentation cases to represent which edges are perturbed in priority. To bridge the gap between theoretical demonstration and practical workflow, we propose EPD, an efficient module to perturb edges based on the priority metric, enabling flexible augmentation or attack.
%Next up, we will operationally unify the workflow of Gaug and Gatk methods via a plug-to-play module to enable flexible adjustments of augmentation and attack.

\normalem
\begin{algorithm}[t]
 \small
  \caption{Edge Priority Detector (Soln. I)} \label{algo:EPDS1}
 $\texttt{INPUT}~$\emph{\textbf{Graph G(V, E), Perturbation Budget b}} \\  
 $\texttt{OUTPUT}~$\emph{\textbf{Augmented and Attacked Edge Sets E$_{aug}$, E$_{atk}$}} \\  
 $Priority~Matrix~I = \{0\}^{|V| \times |V|}$  \\ 
 \textbf{\textit{E}}$_{aug}$, \textbf{\textit{E}}$_{atk}$ = \textbf{\textit{E}} \\
 \For{$v, u \in V~and~v \neq u$}{
    $I_{vu} = 1~\textbf{if}~\emph{\textbf{is\_same\_label}}(v, u)~\textbf{else}~-1$ 
 } 

\While{$Perturbation~Amount < \textbf{b}$}{
 \For{$v, u \in V~and~v \neq u$}{
        \If{$I_{vu} = -1$}{
            \textbf{\textit{E}}$_{atk} \leftarrow ~\emph{\textbf{add\_edge}}(v, u)~\textbf{if}~e_{vu} \notin E_{atk}$ \\
            \textbf{\textit{E}}$_{aug} \leftarrow ~\emph{\textbf{remove\_edge}}(v, u)~\textbf{if}~e_{vu} \in E_{aug}$ \\
        }
        \ElseIf{$I_{vu} = 1$}{
            \textbf{\textit{E}}$_{aug} \leftarrow ~\emph{\textbf{add\_edge}}(v, u)~\textbf{if}~e_{vu} \notin E_{aug}$ \\
            \textbf{\textit{E}}$_{atk} \leftarrow ~\emph{\textbf{remove\_edge}}(v, u)~\textbf{if}~e_{vu} \in E_{atk}$ \\
        }
    }
}     
    %(Stage 1 termination. \textbf{EPD} offline calculates \textbf{\textit{E}}$_{aug}$, \textbf{\textit{E}}$_{atk}$.)

\end{algorithm}

\subsection{Unified Workflow: Edge Priority Detector}

%a module, termed Edge Priority Detector (EPD), which bridges Gaug and Gatk methods up in their workflows. 

This subsection releases the design of EPD, a module that contributes a unified workflow and generates perturbation priorities of edges to establish a quantizable boundary between Gaug and Gatk methods. Based on the perturbation priority metric, we can achieve either effect of augmentation or attack for making flexible perturbations. Perturbation priorities of edges determine which edges will be perturbed with a high probability for augmentation or attack. In practice, prevailing GNNs treat edges in an un-directed graph differentially. A representative example is that GAT \cite{gat} calculates a weight matrix of edges and applies weighted aggregation in node regions, thus highlighting the influences of neighbors corresponding to high-weight edges (i.e., high-priority edges) in aggregating representations. Another example \cite{dfs-gcnjaccard} extends this principle in defending against attacks. In our case, we have previously revealed that the effectiveness of edge perturbation methods comes from their capability of detecting perturbation priorities of edges. To support this argument, we devise a plug-to-play module named EPD and release our solution in Figure \ref{fig:EPD}. Specifically, EPD offers two solutions (Soln. I\&II) to generate effective perturbations. In Solution I, EPD calculates a perturbation priority matrix \textit{I} of edges in a graph. In Solution II, EPD first adopts target-guided modification to alter the adjacency matrix. Targets for attack or augmentation are set as increasing connected components or decreasing the rank of the adjacency matrix, respectively. Then, perturbations will be applied to the modified adjacency matrix, i.e., Adj$^{\prime}$, to generate the priority matrix \textit{I}. Conducting target-guided modification takes extra costs before making perturbations. Thus, one can choose either solution within tolerable costs. We note that both solutions offered by EPD can be done offline to yield a perturbed adjacency matrix. %Next up, we feed it to a GNN for training.

%Importance metrics are critical values to quantify the importance of different parts of data in a fine-grained manner. A representative case of applying importance metrics in deep learning is the attention mechanism \cite{attention}. Typical attention-based techniques capture important (interested) parts of data, such as detecting salient objects in pictures \cite{attention-app1,attention-app2}, which inspires researchers to detect the important connections in graph data for generating attacks \cite{atk-VIKING} or defending attacks \cite{dfs-gcnjaccard}. Our argument has previously revealed that the effectiveness of edge perturbation methods comes from their capability of detecting importance metrics of edges. To support this argument, we devise a plug-to-play module EPD. 

\normalem
\begin{algorithm}[t]
 \small
  \caption{Target-guided Modification: Find Bridge} \label{algo:EPDS2-atk}
 $\texttt{INPUT}~$\emph{\textbf{Graph G(V, E), Adjacency Matrix Adj}} \\  
 $\texttt{OUTPUT}~$\emph{\textbf{Bridge Edge Set E$_{Brg}$}} \\  
 $visited = \{False\}^{1 \times |V|}$  \\ 
 $Time = 0$ \\
\textbf{Function DFS~(\textit{current}, \textit{parent}):} \\
 $visited[current] = True$ \\
 
 $ids[current], low[current] = ++Time $ \\
 \For{$v \in Neighbor\_Set(current)$}{
  \If{$Adj[current][v] = 0 ~ \textbf{OR} ~ v = parent$}{Continue}
  \If{$\textbf{Not} ~ visited[v]$}{
    $\textbf{DFS}~(v, current)$ \\
    $low[current] = \textbf{Min}(low[current], ~low[v])$ \\
    \If{$ids[current] ~<~ low[v]$}{
      $\emph{\textbf{E}}_{Brg} \leftarrow \textbf{\textit{add\_edge}}(current, v)$
      }
    }
    \Else{$low[current] = \textbf{Min}(low[current], ~ids[v])$}
 }
 
\textbf{Function find\_bridge~(\emph{V}):} \\
 \For{$v \in \textbf{V}$}{
    \If{$\textbf{Not} ~ visited[v]$}{$\textbf{DFS}~(v, -1)$}
 }
\end{algorithm}

In Solution I, EPD calculates the perturbation priority matrix \textit{I} based on the graph homophily principle. In real-world graphs, the homophily principle can be exhibited in the node distribution, in which nodes connected in a region tend to have similar attributes or belong to the same class \cite{mcpherson2001birds}. A representative example is that a person generally has many friends sharing the same hobbies, which can describe this ``birds of a feather flock together'' phenomenon well \cite{Networks}. Besides, many existing GNNs, e.g., GCN \cite{gcn} and LGNN \cite{LGNN}, are designed under a strong assumption of homophily \cite{beyond_homo}, and trend to overfit the majority classes, i.e., nodes in homophilic regions. In this case, augmentation and attack can be achieved by changing the homophily of a graph. Specifically, to make augmentation, we enhance the homophily of a node region by removing heterophilic edges and adding homophilic edges between nodes. Literature \cite{CS-GNN} has revealed that neighbors with the same class generally contribute positive information during neighboring aggregation. Therefore, we can make augmentation and facilitate high-quality representation learning by increasing the homophily of multiple node regions in a graph. To inject attacks, we decrease the homophily of a node region by adding heterophilic edges and removing homophilic edges between nodes. An investigation \cite{dfs-gcnjaccard} of graph-related attacks has found that some attack methods connect nodes with different classes to lead to misclassification. Therefore, decreasing the homophily, together with increasing heterophily of node regions in a graph, can generate attacks and further enable effective disturbances for mainstream GNN models. A detailed process can refer to Algorithm \ref{algo:EPDS1}. Note that edges are uniformly added or removed to ensure stochasticity. %To ensure stochasticity, edges are added or removed under a uniform distribution when generating the output edge sets E$_{aug}$, E$_{atk}$.

\begin{figure}[t]
\centering
\includegraphics[width=0.85\columnwidth]{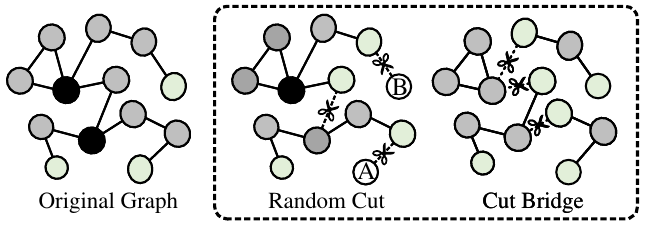}
\caption{Comparisons between naive (random cut) and target-guided (cut bridge) modifications. A node is dyed a darker color if the number of its neighbors is higher.}
\label{fig:CutBridge}
\end{figure}

In Solution II, extra target-guided modifications are conducted to alter the adjacency matrix Adj before calculating the priority matrix \textit{I} via EPD. The modifications are targeted to facilitate the effect of subsequent augmentation or attack. We set different targets for augmentation or attack scenarios according to empirical investigation and experimental evaluation. For attack, we propose to increase the connected components of Adj to increase the number of potential low-degree nodes. Literature \cite{dfs-gcnjaccard,lowdegreeNODES} has previously found that attacks are easier to corrode nodes with fewer neighbors (i.e., low-degree nodes) than nodes with more neighbors. As illustrated in Figure \ref{fig:CutBridge}, randomly modifying a graph may be unreliable and may bring out isolated nodes (e.g., Nodes A, B). In contrast, the target-guided approach is superior in precisely modifying the graph and generating more low-degree nodes. To conduct target-guided modification, we first define an edge $e_{Brg}$ as a bridge if its removal increases the number of connected components of Adj. Thus, removing such bridges can increase the connected components. We propose to find bridges in the given graph based on the Tarjan algorithm \cite{tarjan}. The key idea is to locate bridge edges via Depth First Search (DFS). We give the detailed process in Algorithm \ref{algo:EPDS2-atk}. 
For augmentation, we propose to decrease the rank of Adj based on the conclusion that low-rank graphs provide better generalization capacity and robustness \cite{aug-PTDNet,lowrank1,lowrank2}. Inspired by PTDNet \cite{aug-PTDNet}, we adopt a two-layer MLP to generate weights of edges and calculate the nuclear norm loss for backpropagation. During decreasing the nuclear norm loss, edges associated with minor weights will be removed to decrease the rank of Adj. Different from PTDNet, we remove heterophilic edges in priority and control the number of removed edges per epoch to ensure that a small EPR is produced in the target-guided modification.

\section{Experiments}

In this section, we first demonstrate the effectiveness and flexibility of EPD in making perturbations for augmentation and attack. Then, we conduct extensive experiments to compare EPD and other Gaug/Gatk methods from multiple aspects.

\subsection{Experimental Settings}
\noindent \textit{Model.} We adopt six Gaug and Gatk methods that have released their official repositories. The available code is given in \textit{Table} \ref{tab:resource}. Detailed descriptions can be found in Table \ref{tab:edge_perturbation_summary}. \\
\noindent \textit{Dataset.} We use three real-world graphs to evaluate EPD and other Gaug and Gatk methods. The supported datasets cannot be easily extended to large-scale ones due to the challenges of memory or time overhead \cite{SGA-OOM}. Therefore, we adopt the commonly used datasets of all these methods for evaluation. A summary is given in Table \ref{tab:dataset}. \\ %The split of datasets follows the setting in the literature \cite{fastgcn}.\\
\noindent \textit{Platform.} All experiments are conducted on a Linux server equipped with a 32-core Intel Xeon Platinum 8350C CPU (2.60GHz) and an NVIDIA A100 SXM 80GB.

\begin{table}[t]
\centering
\caption{Summary of used methods and available codes}
\label{tab:resource}
\resizebox{1\columnwidth}{!}{
\begin{tabular}{ccc} \bottomrule Method & Available Code & Commit
\\ \hline
DropEdge\cite{dropedge} &  \href{https://github.com/DropEdge/DropEdge}{https://github.com/DropEdge/DropEdge} &  390cd71\\ \hline
%PTDNet\cite{aug-PTDNet} & \href{https://github.com/flyingdoog/PTDNet}{https://github.com/flyingdoog/PTDNet} & 481c629 \\ \hline
UGS\cite{aug-UGS} & \href{https://github.com/VITA-Group/Unified-LTH-GNN}{https://github.com/VITA-Group/Unified-LTH-GNN} & 4f1b08b \\ \hline
GAUG\cite{aug-GAUG} & \href{https://github.com/zhao-tong/GAug}{https://github.com/zhao-tong/GAug} & 4091bbd \\ \hline
%Metattack\cite{atk-metattack} & \href{https://github.com/danielzuegner/gnn-meta-attack}{https://github.com/danielzuegner/gnn-meta-attack} & 982d900 \\ \hline
%GUA\cite{atk-GUA} & \href{https://github.com/chisam0217/Graph-Universal-Attack}{https://github.com/chisam0217/Graph-Universal-Attack} & 643ce1e \\ \hline
GF-attack\cite{atk-21} & \href{https://github.com/SwiftieH/GFAttack}{https://github.com/SwiftieH/GFAttack} & f84199d \\ \hline
%RL-attack\cite{atk-adversarial} & \href{https://github.com/Hanjun-Dai/graph_adversarial_attack}{https://github.com/Hanjun-Dai/graph\_adversarial\_attack} & f2aaad7 \\
Viking\cite{atk-VIKING} & \href{https://github.com/virresh/viking}{https://github.com/virresh/viking} & 7eb0fde \\ \hline
%Nettack\cite{atk-nettack} & \href{https://github.com/danielzuegner/nettack}{https://github.com/danielzuegner/nettack} & da8ec64 \\ \hline
Topo-attack\cite{atk-topology} & \href{https://github.com/KaidiXu/GCN_ADV_Train}{https://github.com/KaidiXu/GCN\_ADV\_Train} & fc170ba \\ 
%NE-attack\cite{atk-NE-attack} & \href{https://github.com/abojchevski/node_embedding_attack}{https://github.com/abojchevski/node\_embedding\_attack} & dda914c \\ 
%RL-attack\cite{atk-adversarial} & \href{https://github.com/Hanjun-Dai/graph_adversarial_attack}{https://github.com/Hanjun-Dai/graph\_adversarial\_attack} & f2aaad7 \\
\bottomrule
\end{tabular}
}
\end{table}

\begin{table}[ht] 
\centering
\renewcommand\arraystretch{0.9}
\caption{Summary information of the datasets}
\label{tab:dataset}
\begin{tabular}{ccccc} \bottomrule
Dataset & \#Classes & \#Nodes & \#Edges & \#Features \\ \hline
Cora \cite{sen2008collective} & 7 & 2,708 & 5,429 & 1,433\\
Citeseer \cite{sen2008collective} & 6 & 3,327 & 4,732 & 3,703\\
Pubmed \cite{sen2008collective} & 3 & 19,717 & 44,338 & 500\\
%Flickr \cite{dataset:Flickr} & 9 & 7,575 & 239,738 \\
%Cora-ML \cite{dataset:coraml} & 7 & 2,810 & 7,981 \\
%PolBlogs \cite{dataset:PolBlogs} & 2 & 1,222 & 16,714 \\
\bottomrule
\end{tabular}
\end{table}

\subsection{Effectiveness of EPD}
We conduct experiments to evaluate the effectiveness of EPD using widely used GNN variants, including GCN \cite{gcn}, GraphSAGE \cite{graphsage}, SGC \cite{sgc}, and DAGNN \cite{DAGNN}. Like most Gaug and Gatk methods, we only make perturbations on the training set of a graph, leaving the validation and test sets unchanged. In GCN's setting \cite{gcn}, only a small portion of all labels of the training examples will be used in training a model. For example, in Citeseer, connections in the used portion are highly sparse and merely make up 0.38\% of the number of edges in the whole graph. In this case, we adopt the split setting of the datasets used in FastGCN \cite{fastgcn} instead of that of the vanilla GCN \cite{gcn}, i.e., it splits 500 and 1,000 nodes for validation and test, respectively, and uses the rest to train a GNN. To be fair, we retrained all used GNN variants in this split setting to yield the average test accuracy as our baseline. We evaluate the effectiveness of EPD (Soln. I \& II) in making augmentation and attack as follows. Note that configurations of EPD (Soln. I \& II) are open-sourced together with the code for convenient adjustment and reproduction.

\begin{table*}[t]
\centering
\renewcommand\arraystretch{0.8}
\caption{Comparisons of test accuracy between training GNN models with original graphs (the vanilla case) and those with augmented/attacked graphs. We make perturbations via the edge priority matrix generated by EPD (\uline{Soln. I}) to enable augmentation or attack. All results are repeated five times to yield the average ones.}
\label{tab:EPD_Stage1}
\begin{tabular}{cccccccccc}
\bottomrule
\multirow{2}{*}{GNN Model} & \multicolumn{3}{c}{Vanilla Case} & \multicolumn{3}{c}{Augmentation w. EPD} & \multicolumn{3}{c}{Attack w. EPD} \\ \cline{2-10} 
 & \multicolumn{3}{c}{Test Accuracy} & \multicolumn{3}{c}{Test Accuracy} & \multicolumn{3}{c}{Test Accuracy} \\ \hline
 & Cora & Citeseer & Pubmed & Cora & Citeseer & Pubmed & Cora & Citeseer & Pubmed \\ 
GCN & 0.817 & 0.706 & 0.794 & 0.822\tiny$\uparrow$ & 0.721\tiny$\uparrow$ & 0.799\tiny$\uparrow$ & 0.801\tiny$\downarrow$ & 0.695\tiny$\downarrow$ & 0.767\tiny$\downarrow$ \\ 
GraphSAGE & 0.870 & 0.770 & 0.851 & 0.873\tiny$\uparrow$ & 0.774\tiny$\uparrow$ & 0.874\tiny$\uparrow$ & 0.852\tiny$\downarrow$ & 0.768\tiny$\downarrow$ & 0.835\tiny$\downarrow$ \\ 
SGC & 0.864 & 0.768 & 0.842 & 0.866\tiny$\uparrow$ & 0.780\tiny$\uparrow$ & 0.845\tiny$\uparrow$ & 0.857\tiny$\downarrow$ & 0.761\tiny$\downarrow$ & 0.832\tiny$\downarrow$ \\ 
DAGNN & 0.871 & 0.753 & 0.840 & 0.876\tiny$\uparrow$ & 0.764\tiny$\uparrow$ & 0.854\tiny$\uparrow$ & 0.854\tiny$\downarrow$ & 0.738\tiny$\downarrow$ & 0.835\tiny$\downarrow$ \\ 
\bottomrule
\end{tabular}
\end{table*}

\begin{figure*}[t]
\centering
\includegraphics[width=0.99\textwidth]{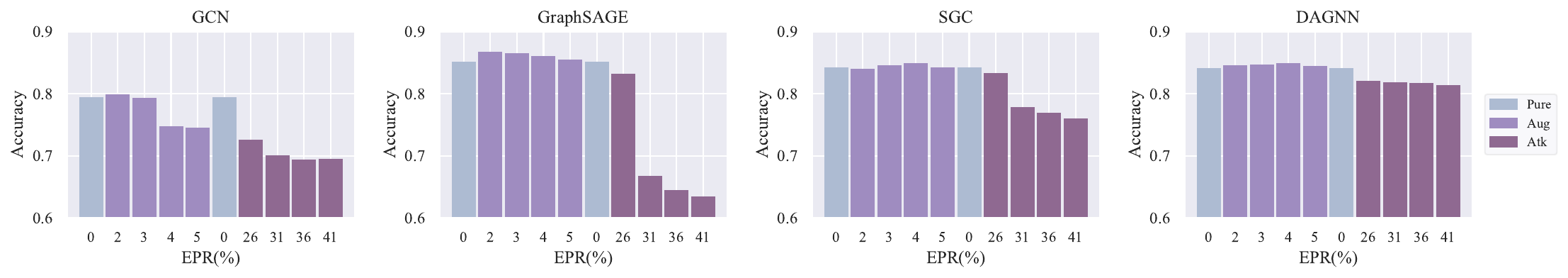}
\caption{Comparisons of test accuracy on Pubmed between four GNN variants under varying EPR. We use EPD (\uline{Soln. II}) to generate the edge priority matrix to make perturbations for augmentation or attack, and increase EPR on Adj$^{\prime}$ before feeding it to GNNs. Note that EPR equaling zero denotes a pure graph without perturbations is fed to GNN to yield the test accuracy.}
\label{fig:EPDS2}
\end{figure*}

\begin{table}[t]
\centering
\renewcommand\arraystretch{0.6}
\caption{Comparisons of test accuracy between the vanilla case and using EPD to inject attack on DAGNN.}
\label{tab:DAGNN}
\begin{tabular}{cccc} \bottomrule
Method & Cora & Citeseer & Pubmed \\ \hline
Vanilla & 0.871 & 0.753 & 0.840 \\
EPD Soln. I  & 0.854\tiny$\downarrow$ & 0.738\tiny$\downarrow$ & 0.835\tiny$\downarrow$ \\
EPD Soln. II & 0.832\tiny$\downarrow$ & 0.641\tiny$\downarrow$ & 0.820\tiny$\downarrow$ \\
\bottomrule
\end{tabular}
\end{table}

As given in Table \ref{tab:EPD_Stage1}, EPD is able to make effective augmentation and attack for all used GNN variants under restricted EPR (e.g., we restrict the EPR of Cora to be 20\% for most GNNs in augmentation cases). As given in Algorithm \ref{algo:EPDS1}, the cost of making perturbations dominates the entire overhead of EPD in Solution I. Therefore, restricting EPR for graphs is a limit on the cost of EPD module, which ensures that effective perturbations generated by EPD are available with minor costs. In Solution II, target-guided modifications toward the graph adjacency matrix are added before calculating the priority matrix \textit{I}. Moreover, EPR is gradually magnified to verify the effectiveness under perturbations of increasing intensity. As illustrated in Figure \ref{fig:EPDS2}, increasing EPR within a certain range can facilitate the effect of attack in most cases. For instance, increasing EPR from 26\% to 31\% brings about 20\% decline in accuracy using GraphSAGE as the backbone. Nevertheless, the facilitation of attacks toward DAGNN is minor when increasing EPR. This is because DAGNN utilizes an adaptive adjustment mechanism to balance the neighboring information and indirectly mitigate the effect of attacks to a certain extent. Similar adaptive strategies are widely used to defend against attacks in GNN domains \cite{Adaptive_dfs_GNN1,Adaptive_dfs_GNN2,Adaptive_dfs_GNN3,Adaptive_dfs_GNN4}. Still, despite the insensitivity of DAGNN to the increasing EPR, we find that Solution II of EPD can offer more effective attacks than Solution I under the same EPR. As given in Table \ref{tab:DAGNN}, The effect of attack is facilitated by increasing connected components of graphs in the target-guided modification. 
To enable augmentation, gradually magnifying ERP can make augmentation and lead to promoted performance in terms of accuracy. Still, we observe the decline of accuracy on some GNNs (e.g., GCN) if EPR exceeds a certain range. We argue that the accuracy of all used GNNs will eventually decline if we magnify EPR continuously. This deduction suits both augmentation and attack cases, i.e., the accuracy under varying EPRs tends to be stable if EPRs exceed their thresholds. %Next, we provide comparisons of accuracy under varying EPR for existing Gaug and Gatk methods in Sec.

\subsection{Comparisons in terms of Test Accuracy and Time}

%Previously, we have found that increasing EPR cannot always bring about promoted effects of augmentation or attack in EPD. The phenomenon also exists in existing Gaug and Gatk methods. In \textbf{Figure} \ref{fig:EXP_EPR}, we give statistics on the accuracy of typical methods under varying EPR and discuss how to select a proper EPR.

This subsection first showcases the comparisons of test accuracy and relative time between EPD and other Gatk/Gaug methods. The recorded time contains processes of making attack/augmentation and training, evaluating the used GNN model. To make fair comparisons in terms of time, we ensure that all selected Gatk and Gaug methods are flexible in adjusting EPR and the training epoch number of GNN. We keep all configurations, including EPR (Cora: 47\%, Citeseer: 37\%, Pubmed: 42\% for attack and 2\% for augmentation), the architecture (a two-layer GCN) and the training epoch number (100) of the used GNN, consistent for these methods on each dataset, and keep method-specific hyperparameters as default. We repeat experiments five times to get the average results. 

\begin{figure*}[t]
\centering
\includegraphics[width=0.9\linewidth]{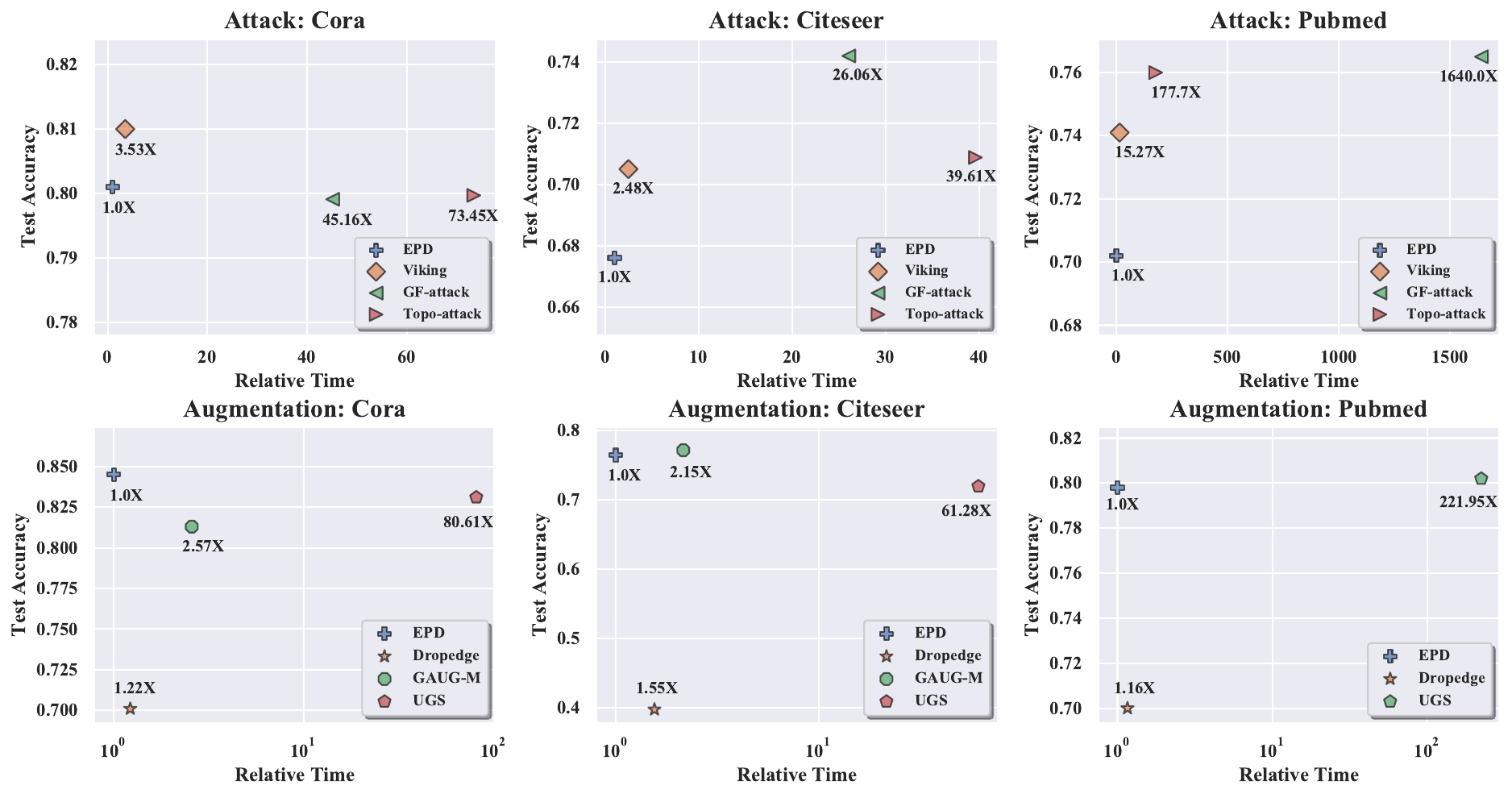}
\caption{Subplots in the 1st row showcase test accuracy and relative time in the attack case while those in the 2nd row showcase that in the augmentation case. Data of GAUG-M \cite{aug-GAUG} is absent on Pubmed as it didn't provide official support for Pubmed.}
\label{fig:EXP_EPD}
\end{figure*}

\begin{figure*}[t]
\centering
\includegraphics[width=0.87\linewidth]{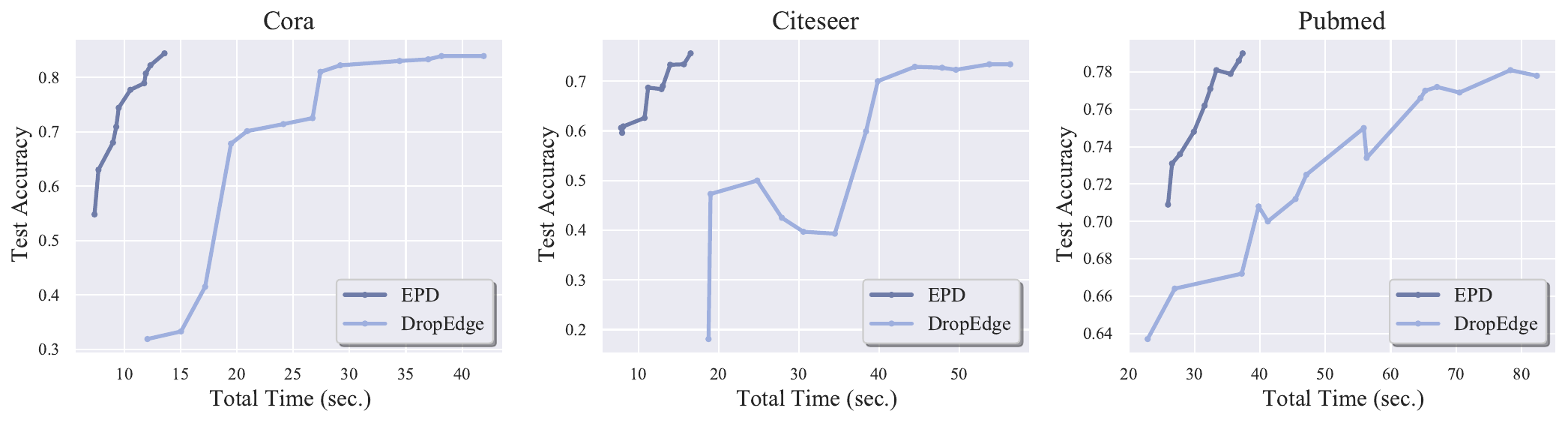}
\caption{Comparisons between EPD and DropEdge in terms of the model convergence.}
\label{fig:EPDvsDropEdge}
\end{figure*}

Results in Figure \ref{fig:EXP_EPD} have demonstrated the superiority of EPD in both efficacy and efficiency. In attack cases, the lower test accuracy, the better. EPD achieves remarkable attack effects on Citeseer and Pubmed, and a comparable effect on Cora. On Pubmed, the time cost of GF-attack and Topo-attack is extremely large (more than 1,000X the cost of EPD), so we have to restrict the perturbation to 200 edges as a compromise. In augmentation cases, the higher test accuracy, the better. EPD achieves desirable accuracy in the shortest time cost on all datasets. The reason for the efficiency of EPD is that it pre-computes the priority metric of a graph once and for all. Then, for attack or augmentation cases and any configurations of EPD (used GNNs, EPR, etc.), the pre-computed metric matrix can be expediently loaded, based on which perturbation will be precisely conducted. The complexity of pre-computing the metric matrix is $\mathcal{O}(N^2)$, where \textit{N} denotes the number of nodes in the training set. We remark that the pre-computation can be offline performed and requires no online resources.

Concerning efficiency, DropEdge is an adequate competitor to EPD in making augmentation since there is only a minor gap in time between DropEdge and EPD. Despite that, EPD is superior to DropEdge in convergence and stability. For a fair comparison, we adopt a plentiful enough training epoch number (around 400) to help make GCN converged when using DropEdge. We run EPD and DropEdge, together with their used GCN, under increasing epoch numbers, and record the corresponding total time and the averaged test accuracy on three datasets. The ``Total Time'' contains processes of making graph data augmentation and execution of GCN. Each point in a curve denotes that a GCN was trained under the ``Total Time'' and evaluated to yield the test accuracy. As shown in \textbf{Figure} \ref{fig:EPDvsDropEdge}, using EPD for augmentation, GCN converges fast to satisfied accuracy, while the same GCN generally takes three times as much time to achieve comparable performance when using DropEdge. We attribute the inferior convergence of DropEdge to the fact that it drops edges and trains GCN with the modified graph in every epoch. In contrast, EPD gets the perturbation done once and for all, then feeds the modified graph to GCN for training. Moreover, randomly dropping edges may lead to unstable and even sub-optimal performance \cite{chen2021structured}, while EPD precisely perturbs edges based on the priority metric. Therefore, EPD is both effective and efficient in making augmentation and attack flexibly.

\vspace{-2mm}
\section{Conclusion}
%This paper aims to answer two open and critical questions related to two categories of edge perturbation methods (Gaug and Gatk) for GNNs, i.e., ``why edge perturbation has a two-faced effect'' and ``what makes edge perturbation flexible and effective''. To this end, from the theoretical perspective, we propose a unified formulation by casting these methods as one optimization problem with different restricted conditions. From the practical perspective, we devise EPD to enable flexible augmentation and attack to bridge these methods up in their workflows. EPD offers a unified workflow and allows tailored adjustments in making augmentation or attack, which inspires bolder attempts to devise novel methods in this domain. Moreover, we have conducted extensive experiments to reveal correlations between the effectiveness of edge perturbation methods and various factors to contribute to the advancement of edge perturbation for augmentation and attack.

This paper aims to answer two open and critical questions related to two categories of edge perturbation methods (Gaug and Gatk) for GNNs, i.e., ``why edge perturbation has a two-faced effect'' and ``what makes edge perturbation flexible and effective''. To this end, from the theoretical perspective, we propose a unified formulation by casting these methods as one optimization problem with different restricted conditions. From the practical perspective, we devise EPD to enable flexible augmentation or attack to bridge these methods up in their workflows. EPD has been experimentally demonstrated as an efficacious and efficient module. It offers a unified workflow and allows tailored adjustments in making augmentation or attack, at the same time, achieves the remarkable performance of augmentation or attack with very short time cost. Moreover, experiments in APPENDIX reveal correlations between the effectiveness of edge perturbation methods and graph intrinsic attributes. We aim to contribute to the advancement of edge perturbation for augmentation and attack and inspire bolder attempts to devise novel methods in this domain.

% or


\section*{Acknowledgment}
This work was supported by National Key Research and Development Program (Grant No. 2023YFB4502305), the National Natural Science Foundation of China (Grant No. 62202451), CAS Project for Young Scientists in Basic Research (Grant No. YSBR-029), and CAS Project for Youth Innovation Promotion Association.

\ifCLASSOPTIONcaptionsoff
  \newpage
\fi

\bibliographystyle{IEEEtran}
\balance
\bibliography{ref}

\begin{figure*}[h]
\centering
\subfigure[Attack Cases: Quantifications of graph attributes of Cora.]{\label{fig6-atk-cr}
\includegraphics[width=0.95\linewidth]{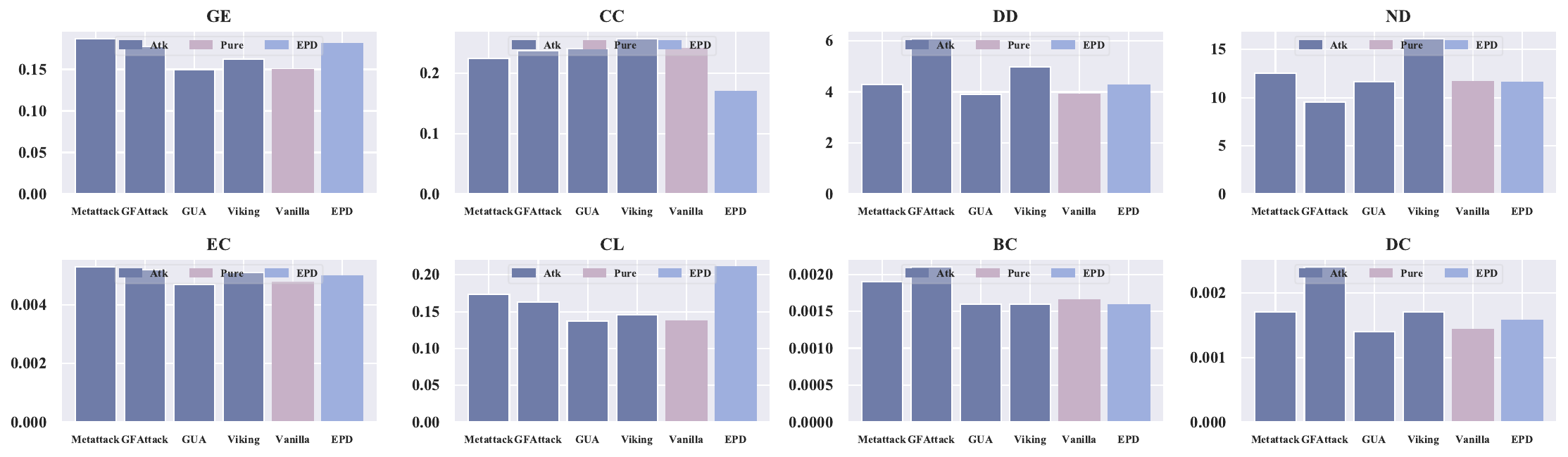}}
%\hspace{0.01\linewidth}
\subfigure[Attack Cases: Quantifications of graph attributes of Citeseer.]{\label{fig6-atk-cs}
\includegraphics[width=0.95\linewidth]{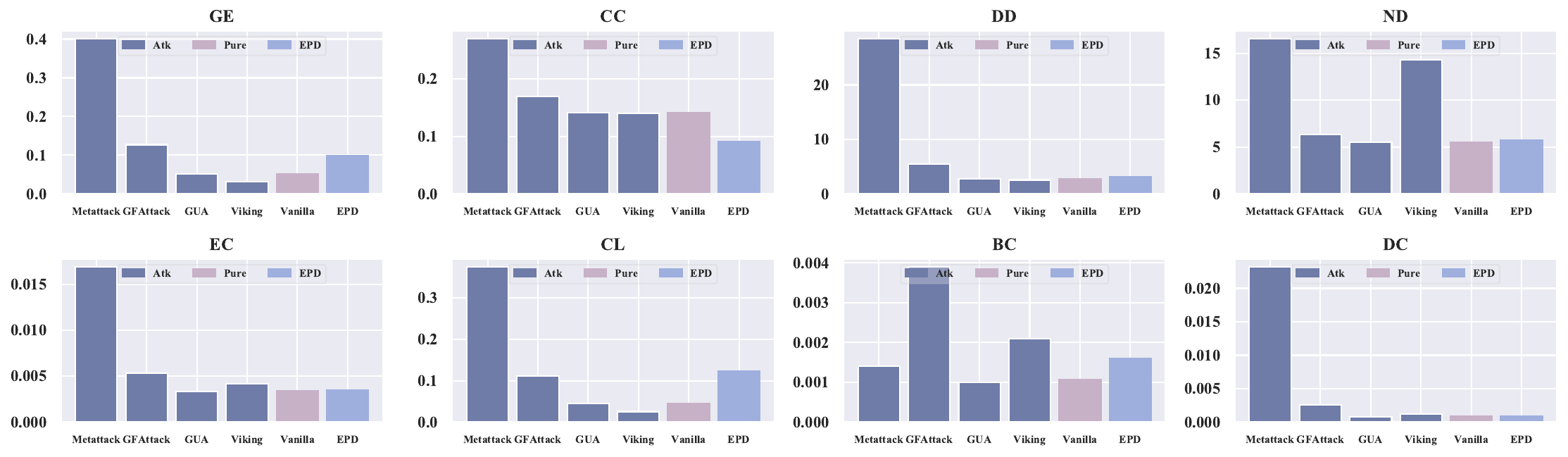}}
%-----------------------------------
\subfigure[Augmentation Cases: Quantifications of graph attributes on Cora.]{\label{fig7-aug-cr}
\includegraphics[width=0.95\linewidth]{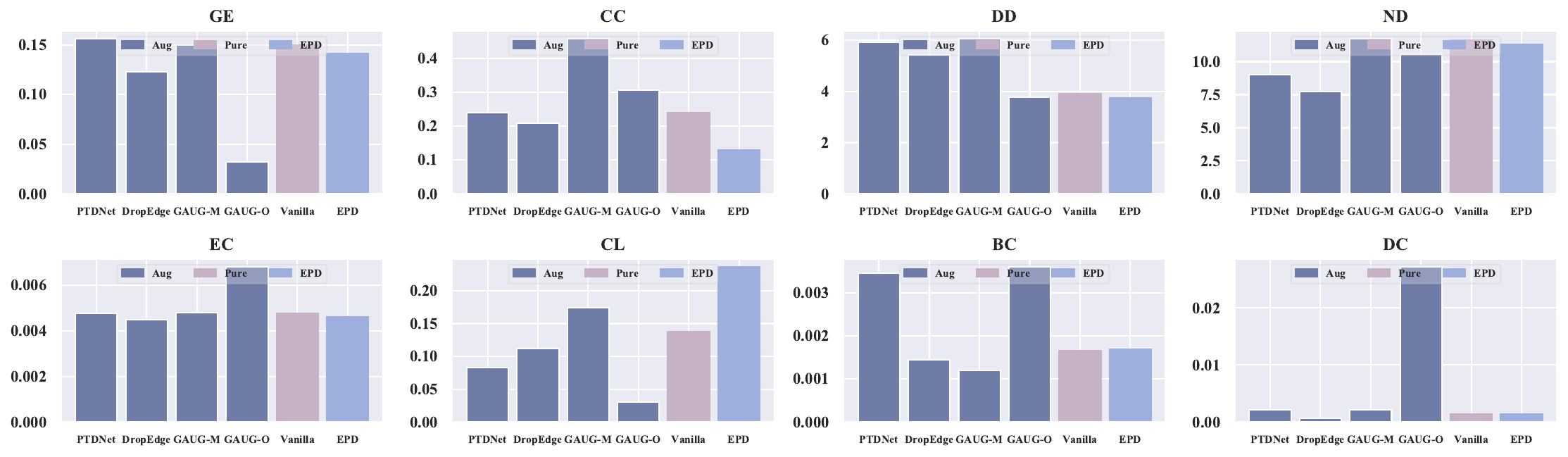}}
\vspace{-3mm}
\subfigure[Augmentation Cases: Quantifications of graph attributes on Citeseer.]{\label{fig7-atk-cs}
\includegraphics[width=0.95\linewidth]{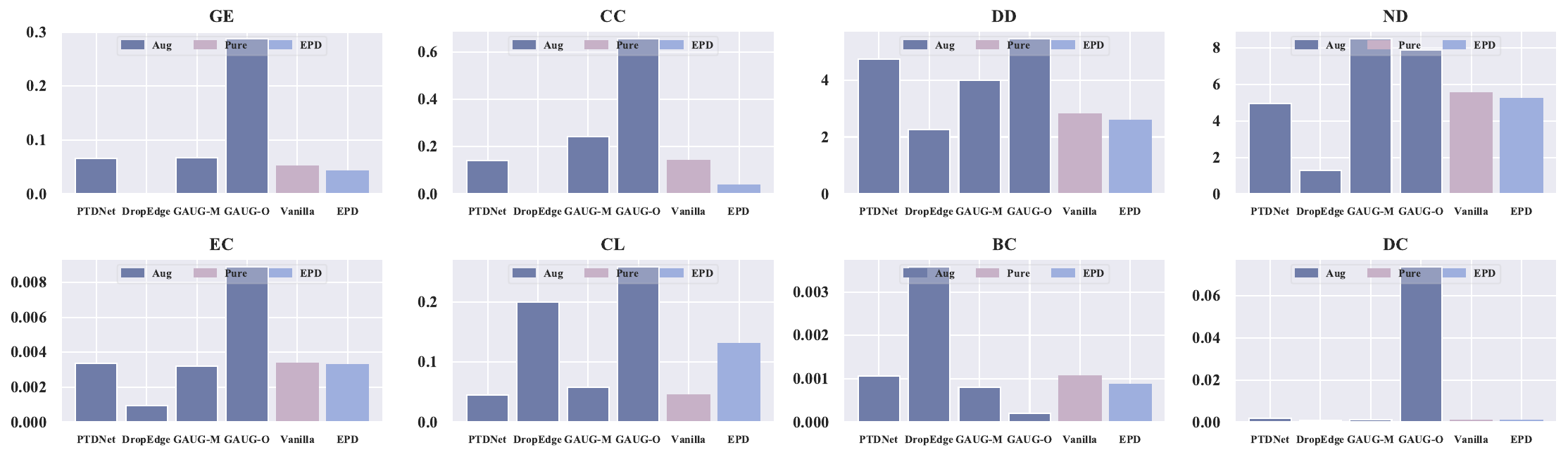}}
\caption{Quantifications of graph attributes of different graphs perturbed by various Gaug and Gatk methods. Note that ``Vanilla'' denotes quantifying graph attributes on a pure graph without any attack/augmentation adopted.}
\label{fig:GraphAttributes}
\end{figure*}

\newpage
\appendix[Comparisons from Graph Analysis Perspective]
This subsection aims to show the impact caused by Gaug and Gatk methods toward graph attributes, which are specific metrics to explain graph characteristics from diverse aspects. Generally, standards to demonstrate the effectiveness of Gaug and Gatk methods are related to the accuracy or deviation in classification/prediction. Quantifications of graph attributes are easily overlooked. Thus, an in-depth analysis should be conducted to reveal the sensitivity of graph attributes to various Gaug/Gatk methods in Figure \ref{fig:GraphAttributes}. We believe quantified results will offer empirical support in devising novel Gaug and Gatk methods. The selected graph attributes are as follows.
\begin{itemize}
    \item Global Efficiency (GE) \cite{atr-GE} denotes the average efficiency over all pairs of nodes in a graph. The efficiency of a pair of nodes is defined as the multiplicative inverse of the shortest path distance between the nodes. 
    \item Clustering Coefficient (CC) \cite{atr-CC} is a metric to measure the cliquishness of a typical neighbor region in a graph.
    \item Degree Distribution (DD) \cite{atr-DD} is a metric to describe the probability distribution of degrees over a graph.
    \item Neighbor Degree (ND) \cite{atr-ND} is the average degree of the neighborhood of each node in a graph. %which implies the connections of local node clusters in extend.
    \item Eigenvector Centrality (EC) \cite{atr-EC} calculates the centrality for a node based on the centrality of its neighbors.
    \item CLoseness Centrality (CL) \cite{atr-CBD} is defined as the reciprocal of the numerical accumulation of the distances between a node and all other nodes in a graph.
    \item Betweenness Centrality (BC) \cite{atr-CBD} denotes the degree to which nodes stand between each other. % a node with higher betweenness centrality would have more control over the network as more information will pass through that node.
    \item Degree Centrality (DC) \cite{atr-CBD} for a node is defined as the fraction of nodes it is connected to. 
\end{itemize}

We adopt EPD, together with eight Gaug and Gatk methods, to make perturbations on Cora and Citeseer, using their official configurations. Note that these metrics are calculated from a global aspect, meaning that a node-level metric is averaged among all nodes to yield a global one for the entire graph. %All methods are executed under their \textbf{official configurations.} %For attack methods, we adopt Metattack, GF-attack, GUA, and Viking to make perturbations. For augmentation methods, we adopt PTDNet, DropEdge, and two variants of GAUG. Next, we analyze the sensitivity of graph attributes to these methods.
%For attack methods, we adopt Metattack \cite{atk-metattack}, GF-attack\cite{atk-21}, GUA \cite{atk-GUA}, and Viking \cite{atk-VIKING} to make perturbations. For augmentation methods, we adopt PTDNet \cite{aug-PTDNet}, DropEdge \cite{dropedge}, and two variants of GAUG \cite{aug-GAUG}. Next, we analyze the sensitivity of graph attributes to these methods.

EPD can make augmentation or inject attack flexibly. In the augmentation case, EPD increases CL and decreases CC. CL is a metric to evaluate the efficiency of information spread among nodes in a graph. A higher value of CL implies an efficient message aggregation during spatial graph convolution. CC is related to the clustering tendency and we attribute the variation of CC to the change of distribution when perturbing edges via EPD. In the attack case, EPD increases GE additionally. GE is a metric used to quantify the concurrent exchange of information across the whole graph. A higher value of GE indicates a higher efficiency of message exchange, which implies that the erroneous information caused by attacks is easily spread in a perturbed graph, thus facilitating the dissemination of erroneous information among attacked nodes.

%\textbf{Metattack} \cite{atk-metattack} increases GE, centrality-related metrics (e.g., EC, CL), and degree-related metrics (e.g., DD, ND, and DC). GE is a metric used to quantify the concurrent exchange of information across the whole graph. A higher value of GE indicates a higher efficiency of message exchange. For Metattack, a high GE implies that the erroneous information caused by attacks is easily spread in a perturbed graph, thus facilitating the dissemination of erroneous information among attacked nodes. Other metrics are related to degree distribution and connected centrality of a graph. These metrics are increased after applying Metattack. This phenomenon indicates that Metattack tends to add poison edges to a graph to enhance the effectiveness of attacks.

Metattack \cite{atk-metattack} increases GE, centrality-related metrics (e.g., EC, CL), and degree-related metrics (e.g., DD, ND, and DC), which are related to degree distribution and connected centrality of a graph. These metrics are increased after applying Metattack. This phenomenon indicates that Metattack tends to add poison edges to a graph to enhance the effectiveness of attacks.

GF-attack \cite{atk-21} mainly increases BC, DD, and DC, despite a minor rise of other attributes after adopting GF-attack compared with the vanilla case. We observe a significant increase in DD and DC with a decline in ND. The observation implies that GF-attack tends to enhance the influence of some central nodes in node regions while decreasing the interactions between the neighbors of these central nodes. Besides, BC, a metric to quantify the intercommunication between different regions in a graph, increases after adopting GF-attack. The phenomenon can be attributed to that more central nodes are produced after perturbing edges. In this context, GF-attack can launch attacks at these central nodes to cause a huge impact on their belonged node regions. 

GUA \cite{atk-GUA} slightly decreases EC and BC. The phenomenon can be attributed to the characteristics of GUA's attack methodology. GUA proposes to perturb edges of anchor nodes of a small number to generate attacks and cause misclassification. We posit that this method is more untraceable and efficacious compared to other Gatk methods. GUA will not significantly modify the topology of graphs, thus yielding trivial effects across all graph attributes.

Viking \cite{atk-VIKING} mainly increases ND and increases DD and BC on Cora and Citeseer, respectively. An increase in ND implies that Viking can make connections in a node region dense. Besides, Viking bears remarkable similarities to GF-attack, except for the ND metric, indicating that Viking also tends to enhance interactions between neighboring nodes.

PTDNet \cite{aug-PTDNet} increases DD and decreases ND on Cora and Citeseer. An increase in DD implies that many nodes with a great many neighbors appear after adopting PDTNet, thus promoting the interactions between node regions. Whereas a decrease in ND may be caused by removing noisy edges since edges between different communities composed by densely connected node regions are regarded as noise and removed from the original graph by PDTNet. 

DropEdge \cite{dropedge} exhibits different effects on Cora and Citeseer. On Cora, all attributes other than DD are decreased, due to the randomly removing edges. On Citeseer, CL and BC are significantly increased, which implies that randomly removing edges results in many isolated node regions. Thus, centrality-related metrics (e.g., CL and BC) are high in these regions. We also find that many metrics on Cora and Citeseer approach zero after adopting DropEdge as promotion of model generalization cannot be achieved until dropping enough edges.

GAUG \cite{aug-GAUG} has two variants: GAUG-M perturbs edges in a graph once and for all, while GAUG-O makes perturbations continuously based on a trainable edge predictor. Overall, DC is significantly increased after adopting GAUG-O since GUAG-O utilizes an edge predictor to add useful edges during training GNNs. On Cora, the graph perturbed by GAUG-O is inferior in connectedness as GE, CC, and CL are numerically lower compared with adopting GAUG-M while rendering an opposite effect on Citeseer. This phenomenon may imply that graphs generated by continuous perturbing edges based on the predicted probability are inconstant in the distribution and centrality, compared with the uniform perturbation.

\end{document}